%% file: eccv2016submission.tex
\documentclass[runningheads]{llncs}
\usepackage{graphicx}
\usepackage{rotating}
\usepackage{wrapfig}
\usepackage{amsmath,amssymb} % define this before the line numbering.
\usepackage{wrapfig}
\usepackage{color}
\usepackage{array}
\usepackage[bookmarks=false]{hyperref}
\hypersetup{
    colorlinks=true,
    linkcolor=blue,
    filecolor=magenta,      
    urlcolor=cyan,
}
\usepackage[width=122mm,left=12mm,paperwidth=146mm,height=193mm,top=12mm,paperheight=217mm]{geometry}
\begin{document}
% \linenumbers

\pagestyle{headings}
\mainmatter

\newcommand{\todo}[1]{\textcolor{red}{TODO: #1}\PackageWarning{TODO:}{#1!}}

%\title{Generating Pixel Motion in Static Scenes} % Replace with your title
\title{An Uncertain Future: Forecasting from Static Images using Variational Autoencoders} % Replace with your title

\author{Jacob Walker, Carl Doersch, Abhinav Gupta, and Martial Hebert}
\institute{The Robotics Institute, Carnegie Mellon University}

\titlerunning{Forecasting from Static Images using Variational Autoencoders}

\authorrunning{J. Walker, C. Doersch, A. Gupta, and M. Hebert}

%def\ECCV16SubNumber{****}  % Insert your submission number here
%\newcommand{\todo}[1]{\textcolor{red}{TODO: #1}\PackageWarning{TODO:}{#1!}}

%\title{Generating Pixel Motion in Static Scenes} % Replace with your title

%\titlerunning{ECCV-16 submission ID \ECCV16SubNumber}

%\authorrunning{ECCV-16 submission ID \ECCV16SubNumber}

%\author{Anonymous ECCV submission}
%\institute{Paper ID \ECCV16SubNumber}

\maketitle

\begin{abstract}
In a given scene, humans can often easily predict a set of immediate
future events that might happen.  
However, generalized pixel-level anticipation in computer vision systems is difficult because machine learning struggles with the ambiguity inherent in predicting the future. 
In this paper, we  focus on predicting the dense trajectory of pixels in a scene --- what will move in the scene, where it will travel, and how it will deform over the course of one second. 
We propose a conditional variational autoencoder as a solution to this problem. 
In this framework, direct inference from the image shapes the distribution of possible trajectories, while latent variables encode any necessary information that is not available in the image. 
We show that our method is able to successfully predict events in a wide variety of scenes and can produce multiple different predictions when the future is ambiguous. 
Our algorithm is trained on thousands of diverse, realistic videos and requires absolutely no human labeling. 
In addition to non-semantic action prediction, we find that our method learns a representation that is applicable to semantic vision tasks. 
\end{abstract}

%%%%%%%%% BODY TEXT
\vspace{-0.2in}
\section{Introduction}
\vspace{-0.05in}
\input{intro.tex}

\vspace{-0.1in}
\section{Background}
\vspace{-0.05in}
\input{related}
\section{Algorithm}
\vspace{-0.05in}
\input{overview}

\vspace{-0.1in}
%\section{Algorithm}
%\vspace{-0.05in}
\input{algorithm}

\vspace{-0.1in}
\section{Experiments}
\vspace{-0.05in}

\input{experiments}

\vspace{-0.1in}
\section{Conclusion}
\vspace{-0.1in}
\input{conclusion}
\vspace{-0.1in}

\bibliographystyle{splncs}
\bibliography{egbib}
\end{document}

%% file: intro.tex
\begin{figure*}[t!]
\centering
\begin{tabular}{ccc}
 \rotatebox{90}{\hspace{0.13in}{\bf{Prediction 1}}} &
 \includegraphics[height=0.14\textheight,width=0.3\textwidth]{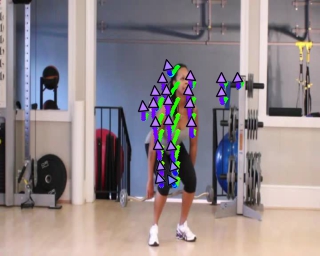} &
\includegraphics[height=0.14\textheight,width=0.6\textwidth]{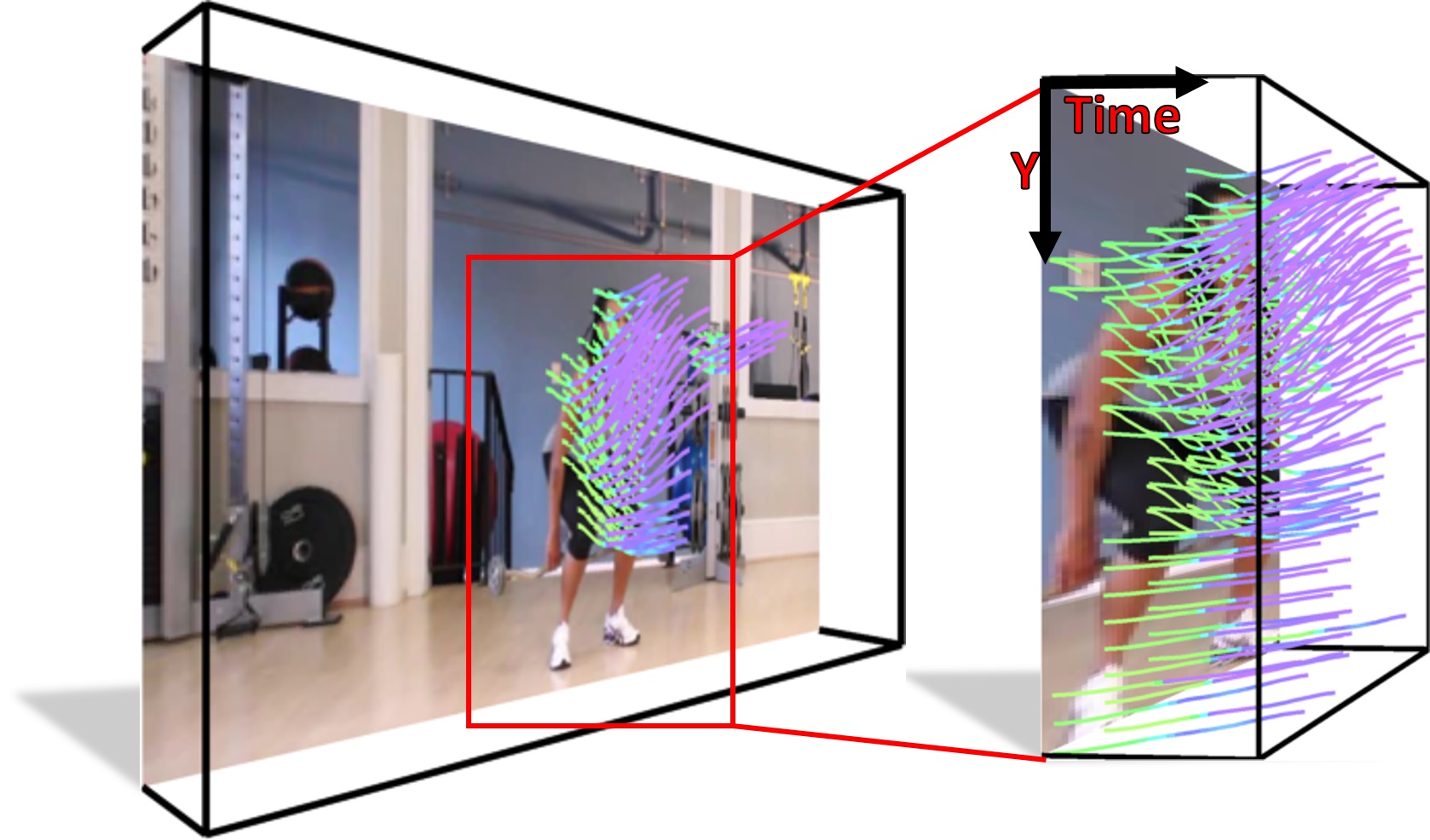} \\
 \rotatebox{90}{\hspace{0.13in}{\bf{Prediction 2}}} &
\includegraphics[height=0.14\textheight,width=0.3\textwidth]{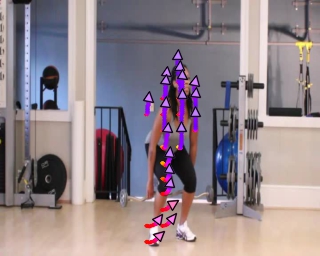} &
\includegraphics[height=0.14\textheight,width=0.6\textwidth]{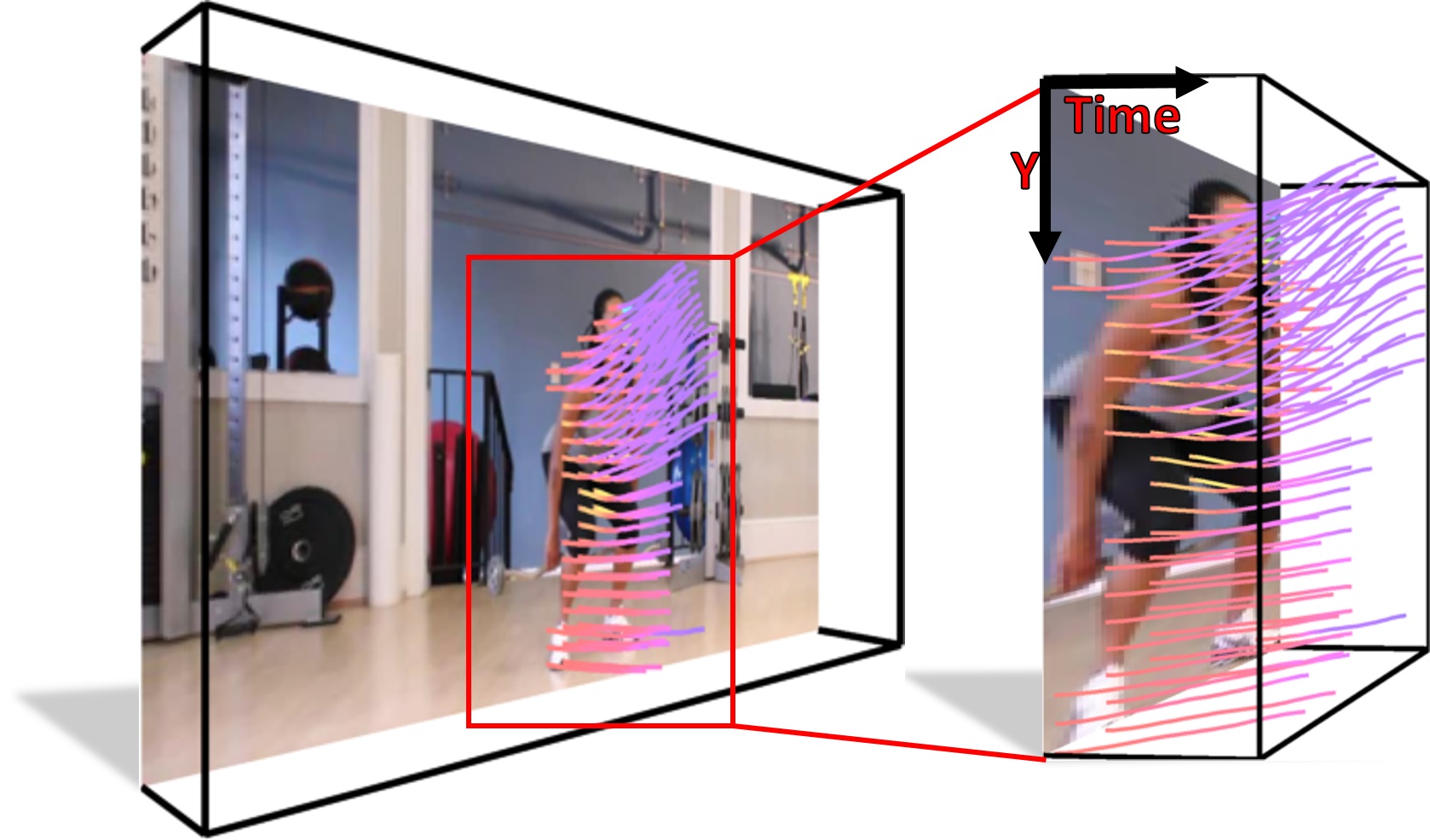} \\
 & (a) Trajectories on Image & (b) Trajectories in Space-Time
\setlength{\tabcolsep}{10pt}
\end{tabular}{}
\begin{tabular}{ p{10cm} r }
\vspace{-0.2in}
\caption{\footnotesize Consider this picture of a woman in the gym --- she could move up or down. Our framework is able to predict multiple correct one-second motion trajectories given the scene. The directions of the trajectories at each point in time are color-coded according to the square on the right. On the left is the projection of the trajectories on the image plane. The right diagram shows the complexity of the predicted motions in space time. Best seen in our \href{http://www.cs.cmu.edu/~jcwalker/DTP/DTP.html}{videos}.}
\label{teaser} & \raisebox{-1.0\height}{\protect\includegraphics[scale=0.70]{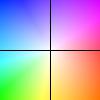}}
\end{tabular}{}
\vspace{-0.45in}
\end{figure*}

Visual prediction is one of the most fundamental and difficult tasks in computer vision. For example, consider the woman in the gym in Figure~\ref{teaser}. 
We as humans, given the context of the scene and her sitting pose, know that she is probably performing squat exercises. 
However, going beyond the action label and predicting the future leads to multiple, richer possibilities. 
The woman might be on her way up and will continue to go up, or she might be on the way down and continue to descend further. 
Those motion trajectories might not be exactly vertical, as the woman might lean or move her arms back as she ascends. 
%and might have some lateral movements etc. 
While there are multiple possibilities, the space of possible futures is significantly smaller than the space of all possible visual motions. 
For example, we know she is not going to walk forward, she is not going to perform an incoherent action such as a head-bob, and that her torso will likely remain in one piece. 
In this paper, we propose to develop a generative framework which, given a static input image, outputs the space of possible future actions. 
The key here is that our model characterizes the whole distribution of future states and can be used to sample multiple possible future events.

Even if we acknowledge that our algorithm must produce a distribution over many possible predictions, it remains unclear what is the output space of futures the algorithm should be capable of predicting. An ideal algorithm would predict everything that might be relevant to a human or robot interacting with the scene, but this is far too complicated to be feasible with current methods.
A more tractable approach is to predict dense trajectories~\cite{Yuen10}, which are simpler than pixels but still capture most of a video's content.  
While this representation is intuitive, the output space is high dimensional and hard to parametrize, an issue which forced~\cite{Yuen10} to use a Nearest Neighbor algorithm and transfer raw trajectories. Unsurprisingly, the algorithm is computationally expensive and fails on testing images which do not have globally similar training images.  
Many approaches try to simplify the problem, either by using some semantic form of prediction~\cite{Savarese14}, predicting agent-based trajectories in a restricted domain~\cite{Walker14}, or just predicting the optical flow to the next frame~\cite{Walker15,Pintea14}. 
However, each of these representations compromise the richness of output in either the spatial domain (agent-based), the temporal domain (optical flow), or both (semantic). 
Therefore, the vision community has recently pushed back and directly attacked the problem of full blown visual prediction: recent works have proposed predicting pixels~\cite{Srivastava15,Ranzato14} or the high dimensional fc7 features~\cite{Vondrick15} themselves. 
However, these approach suffer from a number of drawbacks.  Notably, the output space is high dimensional and it is difficult to encode constraints on the output space, e.g., pixels can change colors every frame. There is also an averaging effect of multiple possible predictions which leads to blurry predictions.

In this paper, we propose to address these challenges. 
We propose to revisit the idea of predicting dense trajectories at each and every pixel using a feed-forward Convolutional Network. 
Using dense trajectories restricts the output space dramatically which allows our algorithm to learn robust models for visual prediction with the available data. 
However, the dense trajectories are still high-dimensional and the output still has multiple modes. 
In order to tackle these challenges, we propose to use variational autoencoders to learn a low-dimensional latent representation of the output space conditioned on an input image. 
Specifically, given a single frame as input, our \textit{conditional} variational auto-encoder outputs a mapping from noise variables---sampled from a normal distribution $\mathcal{N}(0,1)$---to output trajectories at every pixel.  
Thus, we can naively sample values of the latent variables and pass them through the mapping in order to sample different predicted trajectories from the inferred conditional distribution.
%Specifically, we learn \textit{conditional} variational auto-encoders which, given a feature representation of the first frame and latent values sampled randomly from a normal distribution, produce trajectories at every pixel. 
Unlike other applications of variational autoencoders that generate outputs a priori~\cite{Kingma14a,Kingma14b,Gregor15}, we focus on generating them \textit{given the image}. 
Conditioning on the image is a form of inference, restricting the possible motions based on object location and scene context. 
Sampling latent variables during test time then allows us to explore the space of possible actions in the given scene. %Also, note that for the same input and a different sample of latent variable, we get a completely different set of output trajectories.

\noindent {\bf Contributions:}
Our paper makes three contributions. First, we demonstrate that prediction of dense pixel trajectories is a plausible approach to general, non-semantic, self-supervised visual prediction. 
Second, we propose a conditional variational auto-encoder as a solution to this problem, a model that performs inference on an image by conditioning the distribution of possible movements on a scene. 
Third, we show that our model is capable of learning representations for various recognition tasks with less data than conventional approaches.

%% file: related.tex
There have been two main thrusts in recent years concerning visual activity forecasting.
The first is an unsupervised, largely non-semantic approach driven by large amounts of data. 
These methods often focus on predicting low level features such as pixels or the motion of pixels. 
One early approach used nearest-neighbors, making predictions for an image by matching it to frames in a large collection of videos and transferring the associated motion tracks~\cite{Yuen10}.  
An improvement to this approach used dense-SIFT correspondence to align the matched objects in two images~\cite{Liu11}.  
This form of nearest-neighbors, however, relied on finding global matches to an image's entire contents.  
One way this limitation has been addressed is by breaking the images into smaller pieces based on mid-level discriminative patches~\cite{Walker14}. 
Another way is to treat prediction as a regression problem and use standard machine learning, but existing algorithms struggle to capture the complexity of real videos.  
Some works simplify the problem to predicting optical flow between pairs of frames~\cite{Pintea14}.  
Recently, more powerful deep learning approaches have improved on these results~\cite{Walker15}, and some works even suggest that it may be possible to use deep networks to predict raw pixels~\cite{Srivastava15,Ranzato14}.  However, even deep networks still struggle with underfitting in this scenario. 
Hence,~\cite{Vondrick15} considered predicting the top-level CNN features of future frames rather than pixels. 

Since predicting low level features such as pixels is often so difficult, other works have focused on predicting more semantic information. 
For example, some works break video sequences into discrete actions, and attempt to detect the earliest frames in an action in order to predict the rest~\cite{Hoai12}.  
Others predict labeled  human walking trajectories~\cite{Kitani12} or object trajectories~\cite{Mottaghi16} in restricted domains. 
Finally, supervised learning has also been used to model and forecast labeled human-human~\cite{Savarese14,Kitani14}, and human-object~\cite{Koppula13,Fouhey14} interactions.  

A key contribution of our approach is that we explicitly model a distribution over possible futures in the high-dimensional, continuous output space of trajectories. 
That is, we build a generative model over trajectories given an image, and we rely on the recent generative model framework of variational autoencoders (VAEs) to solve the problem. 
VAEs have already shown promise in a number of domains involving generating pixels, including handwritten digits~\cite{Kingma14a,Salimans15}, faces~\cite{Kingma14a,Rezende14}, house numbers~\cite{Kingma14b}, CIFAR images~\cite{Gregor15}, and even face pose~\cite{Kulkarni15}. 
Our work shows that VAEs extend to the novel domain of motion prediction in the form of trajectories.

Our approach has multiple advantages over previous works. First, our approach requires no human labeling. While~\cite{Mottaghi16} also predicted long-term motion of objects, it required manual labels. Second, our approach is able to predict for a relatively long period of time: one second. While~\cite{Pintea14,Walker15} needed no human labeling, they only focused on predicting motion for the next instant frame.  While~\cite{Walker15} did consider long-term optical flow as a proof of concept, they did not tackle the possibility of multiple potential futures.  Finally, our algorithm predicts from a single image---which may enable graphics applications that involve animating still photographs---while many earlier works require video inputs~\cite{Srivastava15,Ranzato14}.
%, not images, and lead to blurring when applied to datasets such as the UCF-101~\cite{Soomro12}. Our method uses only one image and is able to handle multi-domain datasets. 

%In an effort to move towards parameteric approaches, as a next step there was an effort to use semantic form of visual prediction~\cite{Hoai12}. However, semantic prediction fails to provide details of exactly how different elements in the scene are going to move. Other approaches tried to use an agent-based framework in restrictive domains. Specifically, these approaches use a restricted domain such as aerial images and perform prediction as path-planning and represent the future in terms of trajectories of agents~\cite{Kitani12,Walker14}. However, such approaches are domain restricted and do not generalize to generic videos.

%% file: overview.tex
We aim to predict the motion trajectory for each and every pixel in a static, RGB image over the course of one second.  
Let $X$ be the image, and $Y$ be the full set of trajectories. The raw output space for $Y$ is very large---over four million dimensions assuming a 320x240 image at 30 fps---and it is continuous.
We can simplify the output space somewhat by encoding the trajectories in the frequency spectrum in order to reduce dimensionality. However, a more important difficulty than raw data size is that the output space is not unimodal; an image may have multiple reasonable futures. 
\vspace{-0.1in}
\subsection{Model}
\vspace{-0.05in}

A simple regressor---even a deep network with millions of parameters---will struggle with predicting one-second motion in a single image as there may be many plausible outputs.  
Our architecture augments the simple regression model by adding another input $z$ to the regressor (shown in Figure~\ref{architecture}(a)), which can account for the ambiguity.  
At test time, $z$ is random Gaussian noise: passing an image as input and sampling from the noise variable allows us to sample from the model's posterior given the image.  
That is, if there are multiple possible futures given an image, then for each possible future, there will be a different set of $z$ values which map to that future.
Furthermore, the likelihood of sampling each possible future will be proportional to the likelihood of sampling a $z$ value that maps to it.
Note that we assume that the regressor---in our case, a deep neural network---is capable of encoding dependencies between the output trajectories.  
In practice, this means that if two pixels need to move together even if the direction of motion is uncertain, then they can simply be influenced by the same dimension of the $z$ vector.
\vspace{-0.1in}
\subsection{Training by ``Autoencoding''}
\vspace{-0.05in}
It is straightforward to sample from the posterior at test time, but it is much less straightforward to train a model like this.
The problem is that given some ground-truth trajectory $Y$, we cannot directly measure the probability of the trajectory given an image $X$ under a given model; this prevents us from performing gradient descent on this likelihood.
It is in theory possible to estimate this conditional likelihood by sampling a large number of $z$ values and constructing a Parzen window estimate using the resulting trajectories, but this approach by itself is too costly to be useful.

\begin{figure*}[t!]
\centering
\begin{tabular}{c||c}
 \includegraphics[trim = 2.0in 0.8in 2.0in 0.8in,clip,width=0.5\textwidth,height=0.20\textheight]{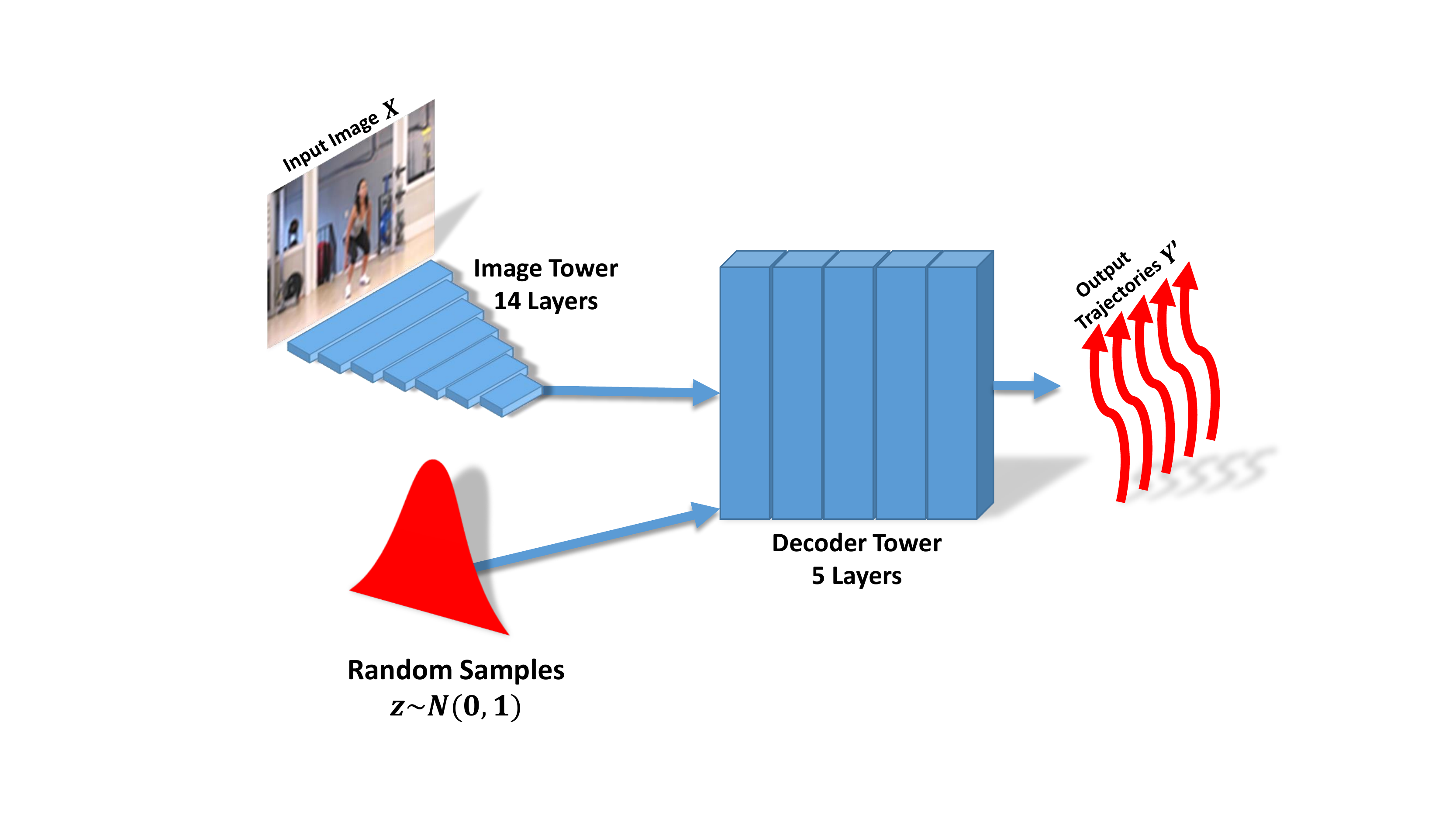} &
 \includegraphics[trim = 1.5in 0.5in 1.5in 0.5in,clip,width=0.5\textwidth,height=0.20\textheight]{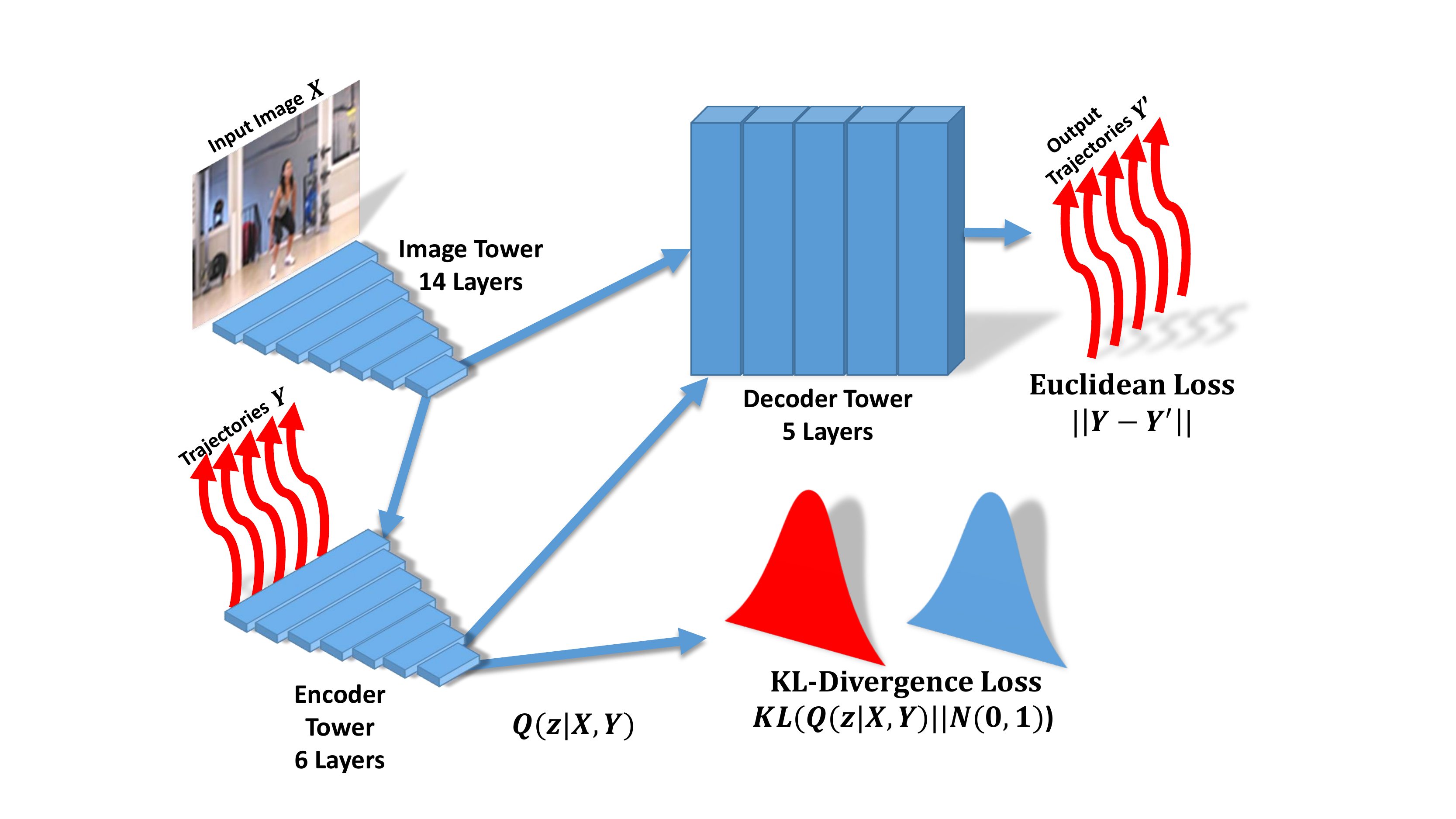}\\
{\footnotesize (a) Testing Architecture} & {\footnotesize (b) Training Architecture} \\
\end{tabular}
\caption{Overview of the architecture. During training, the inputs to the network include both the image and the ground truth
trajectories. A variational autoencoder encodes the joint image and trajectory space, while the decoder produces trajectories depending both on the image information as well as output from the encoder. During test time, the only inputs to the decoder are the image and latent variables sampled from a normal distribution.  }
\vspace{-0.2in}
\label{architecture}
\end{figure*}

Variational Autoencoders~\cite{Kingma14a,Kingma14b,Gregor15,Doersch16} modify this approach and make it tractable.
The key insight is that the vast majority of samples $z$ contribute almost nothing to the overall likelihood of $Y$.
Hence, we should instead focus only on those values of $z$ that are likely to produce values close to $Y$.
We do this by adding another pathway $Q$, as shown in Figure~\ref{architecture}(b), which is trained to map the output $Y$ to the values of $z$ which are likely to produce them.
That is, $Q$ is trained to ``encode'' $Y$ into the latent $z$ space such that the values can be ``decoded'' back to the trajectories.
The entire pipeline can be trained end-to-end using reconstruction error.
An immediate objection one might raise is that this is essentially ``cheating'' at training time: the model sees the values that it is trying to predict, and may just copy them to the output.
To prevent the model from simply copying, we force the encoding to be lossy. 
The $Q$ pathway does not produce a single $z$, but instead, produces a distribution over $z$ values, which we sample from before decoding the trajectories.  
We then directly penalize the information content in this distribution, by penalizing the $\mathcal{KL}$-divergence between the distribution produced by $Q$ and the trajectory-agnostic $\mathcal{N}(0,1)$ distribution.  
The model is thereby encouraged to extract as much information as possible from the input image before relying on encoding the trajectories themselves.
Surprisingly, this formulation is a very close approximation to maximizing the posterior likelihood $P(Y|X)$ that we are interested in.
In fact, if our encoder pathway $Q$ can estimate the exact distribution of $z$'s that are likely to generate $Y$, then the approximation is exact.
%Our method accounts for this ambiguity by adding an autoencoder to the network that encodes latent variables that cannot be inferred from the image. Our fully-convolutional architecture, shown in Figure ~\ref{architecture}, consists of a 14-layer image tower that receives only input from the image, a 6-layer encoder of the joint trajectory-image space, and a 5-layer decoder that receives input from the image tower and the autoencoder. The loss function consists of two components --- the Euclidean loss between the predicted trajectories $Y'$ and the ground truth trajectories $Y$ --- and the $\mathcal{KL}$-divergence loss between the distribution of the encoded latent variables $q(z|X,Y)$ and the normal distribution $\mathcal{N}(0,1)$. 

%The trajectories are also normalized by the average magnitude in the image, and therefore the network also outputs the estimated average global magnitude. During test time, only the image tower and the decoder are used. The decoder is fed latent variables sampled from a normal distribution, and based on the input image the decoder will output a trajectories that are relevant for the context of the scene. 

%% file: algorithm.tex
\vspace{-0.1in}
\subsection{The Conditional Variational Autoencoder}
\vspace{-0.05in}
%Our goal is to construct a distribution $P(Y|X)$ over trajectories $Y$ given an image $X$. 
We now show mathematically how to perform gradient descent on our conditional VAE.  We first formalize the model in Figure~\ref{architecture}(a) with the following formula:
%distribution using a pair of deterministic functions:
\vspace{-0.05in}
\begin{equation}\label{eq:model}
    Y=\mu(X,z)+\epsilon% \circ \sigma(X,z)
\end{equation}
\vspace{-0.05in}
\noindent where $z \sim \mathcal{N}(0,1)$, $\epsilon \sim \mathcal{N}(0,1)$ are both white Gaussian noise. %, and $\circ$ denotes an elementwise product.  
%Note that all dependencies between the trajectories $Y$ at different locations must be represented in $\mu$ and $\sigma$: for example two dimensions of $Y$ can be made to correlate strongly if they depend primarily on the same dimension of $z$.  
We assume $\mu$ is implemented as a neural network. %, which are capable of encoding all dependencies in the output trajectories.

Given a training example $(X_i,Y_i)$, it is difficult to directly infer $P(Y_i|X_i)$ without sampling a large number of $z$ values.  %, since the space of $z$ variables that are likely to produce $Y_i$ is often very small.  
Hence, the variational ``autoencoder'' framework first samples $z$ from some distribution different from $\mathcal{N}(0,1)$ (specifically, a distribution of $z$ values which are likely to give rise to $Y_i$ given $X_i$), and uses that sample to approximate $P(Y|X)$ in the following way.  
Say that $z$ is sampled from an arbitrary distribution $z \sim Q$ with p.d.f. $Q(z)$.  
By Bayes rule, we have:
\vspace{-0.05in}
\begin{equation}
    E_{z\sim Q}\left[\log P(Y_i|z,X_i)\right]=E_{z\sim Q}\left[\log P(z|Y_i,X_i) - \log P(z|X_i) + \log P(Y_i|X_i) \right]
\end{equation}
\vspace{-0.05in}
\noindent Rearranging the terms and subtracting $E_{z\sim Q}\log Q(z)$ from both sides:
\vspace{-0.05in}
\begin{equation}
\begin{array}{c}
  \log P(Y_i|X_i) - E_{z\sim Q}\left[\log Q(z)-\log P(z|X_i,Y_i)\right]=\hspace{10em}\\
  \hspace{10em}E_{z\sim Q}\left[\log P(Y_i|z,X_i)+\log P(z|X_i)-\log Q(z)\right]
\end{array}
\end{equation}
\vspace{-0.05in}
\noindent Note that $X_i$ and $Y_i$ are fixed, and $Q$ is an arbitrary distribution.  
Hence, during training, it makes sense to choose a $Q$ which will make $E_{z\sim Q}[\log Q(z)-$ 
$\log P(z|X_i,Y_i)]$ (a $\mathcal{KL}$-divergence) small, such that the right hand side is a close approximation to $\log P(Y_i|X_i)$.  
Specifically, we set $Q = \mathcal{N}(\mu^{\prime}(X_i,Y_i),$ $\sigma^{\prime}(X_i,Y_i))$ for functions $\mu^{\prime}$ and $\sigma^{\prime}$, which are also implemented as neural networks, and which are trained alongside $\mu$.  
We denote this p.d.f. as $Q(z|X_i,Y_i)$.  We can rewrite some of the above expectations as $\mathcal{KL}$-divergences to obtain the standard variational equality:
\vspace{-0.05in}
\begin{equation}
\begin{array}{c}
  \log P(Y_i|X_i) - \mathcal{KL}\left[Q(z|X_i,Y_i)\| P(z|X_i,Y_i)\right]=\hspace{10em}\\
  \hspace{10em}E_{z\sim Q}\left[\log P(Y_i|z,X_i)\right]-\mathcal{KL}\left[Q(z|X_i,Y_i)\| P(z|X_i)\right]
\end{array}
\end{equation}
\vspace{-0.05in}
We compute the expected gradient with respect to only the right hand side of this equation, so that we can perform gradient ascent and maximize both sides.  
Note that this means our algorithm is accomplishing two things simultaneously: it is maximizing the likelihood of $Y$ while also training $Q$ to approximate $P(z|X_i,Y_i)$ as well as possible.  
Assuming a high capacity $Q$ which can accurately model $P(z|X_i,Y_i)$, this second $\mathcal{KL}$-divergence term will tend to 0, meaning that we will be directly optimizing the likelihood of $Y$.
To perform the optimization, first note that our model in Equation~\ref{eq:model} assumes $P(z|X_i) = \mathcal{N}(0,1)$, i.e., $z$ is independent of $X$ if $Y$ is unknown.  
Hence, the $\mathcal{KL}$-divergence may be computed using a closed form expression, which is differentiable with respect to the parameters of $\mu^{\prime}$ and $\sigma^{\prime}$.  
We can approximate the expected gradient of $\log P(Y_i|z,X_i)$ by sampling values of $z$ from $Q$.  
The main difficulty, however, is that the distribution of $z$ depends on the parameters of $\mu^{\prime}$ and $\sigma^{\prime}$, which means we must backprop through the apparently non-differentiable sampling step.
We use the ``reparameterization trick''~\cite{Rezende14,Kingma14a} to make sampling differentiable.
Specfically, we set $z_i=\mu^{\prime}(X_{i},Y_{i})+\eta \circ \sigma^{\prime}(X_{i},Y_{i})$, where $\eta\sim\mathcal{N}(0,1)$ and $\circ$ denotes an elementwise product.  This makes $z_i\sim Q$ while allowing the expression for $z_i$ to be differentiable with respect to $\mu^{\prime}$ and $\sigma^{\prime}$.

%Note that if $X_i$ is the same for  
%Regarding our architecture, the five layers of the encoder tower ultimately estimate $Q(z|X_i,Y_i)$, taking as inputs image $X_i$ and trajectories $Y_i$. $\mu^{\prime}$ and $\sigma^{\prime}$ are directly based on features average-pooled over the entire fifth layer. In this way, the latent variables are unable to depend on location-specific information. The resulting latent variables $z_i$ are then passed through convolutional layers that multiply and subtract from the top-most layer from the image tower. These operations condition the decoder on the image information. Finally, the top-most merged tower -- consisting of five layers --- is a decoder of the $z_i$ approximating $ P(Y_i|X_i)$. 
\vspace{-0.1in}
\subsection{Architecture} 
\vspace{-0.05in}
Our conditional variational autoencoder requires neural networks to compute three separate functions: $\mu(X,z)$ which comprises the ``decoder'' distribution of trajectories given images ($P(Y|X,z)$), and $\mu^{\prime}$ and $\sigma^{\prime}$ which comprise the ``encoder'' distribution ($Q(z|X,Y)$).  However, much of the computation can be shared between these functions: all three depend on the image information, and both $\mu^{\prime}$ and $\sigma^{\prime}$ rely on exactly the same information (image and trajectories).  Hence, we can share computation between them.  The resulting network can be summerized as three ``towers'' of neural network layers, as shown in Figure~\ref{architecture}.  First, the ``image'' tower processes each image, and is used to compute all three quantities.  Second is the ``encoder'' tower, which takes input from the ``image'' tower as well as the raw trajectories, and has two tops, one for $\mu^{\prime}$ and one for $\sigma^{\prime}$, which implements the $Q$ distribution.  This tower is discarded at test time.  Third is the ``decoder'' tower, which takes input from the ``image'' tower as well as the samples for $z$, either produced by the ``encoder'' tower (training time) or random noise (test time).  All towers are fully-convolutional.  The remainder of this section details the design of these three towers.
%Hence, we implement $\mu^{\prime}$ and $\sigma^{prime}$ via a single neural network which has two ``tops:'' one for $\mu^{\prime}$ and another for $\sigma^{prime}$.
%Our network is fully convolutional and consists of two towers --- an image tower and an autoencoder --- that merge into one top tower, the decoder, for the final outputs. We describe the details of each component of our network below.  

\noindent {\bf Image Tower:}
The first, the image data tower, receives only the 320x240 image as input. The first five layers of the image tower are almost identical to the traditional AlexNet~\cite{Krizhevsky12} architecture with the exception of extra padding in the first layer (to ensure that the feature maps remain aligned to the pixels).%are divisible by the size of the spatial output. 
We remove the fully connected layers, since we want the network to generalize across translations of the moving object.  
We found, however, that 5 convolutional layers is too little capacity, and furthermore limits each unit's receptive field to too small a region of the input.  
Hence, we add nine additional 256-channel convolutional layers with local receptive fields of 3. 
%While networks such as VGG-net ~\cite{Simonyan14} and GoogLeNet ~\cite{Szegedy15} are established architectures with high capacity, our architecture results in comparatively less computation time. 
%In addition, all convolutional layers are batch-normalized. 
To simplify notation, denote $C(k,s)$ as a convolutional layer with kernel size $k$ and receptive field size $s$. Denote $LRN$ as a Local Response Normalization, and $P$ as a max-pooling layer. Let $\rightarrow C(k,s)_{i} \rightarrow C(k,s)_{i+1}$ denote a series of stacked convolutional layers with the same kernel size and receptive field size. 
This results in a network described as: $C(96,11) \rightarrow LRN \rightarrow P \rightarrow C(256, 5) \rightarrow LRN \rightarrow P \rightarrow C(384,3) \rightarrow C(384,3) \rightarrow C(256,3)_{1} \rightarrow C(256,3)_{2} ... \rightarrow C(256,3)_{10}.$

\noindent {\bf Encoder Tower:}
%The second tower is a variational auto-encoder of latent variables. 
We begin with the frequency-domain trajectories as input, and downsample them spatially such that they can be concatenated with the output of the image tower.  
The encoder tower takes this tensor as input and processes them with five convolutional layers similar to AlexNet, although the input consists of output from the image tower and trajectory data concatenated into one input data layer. 
%The convolutional layers in the encoder tower are also subject to batch normalization. 
After the fifth layer, two additional convolutional layers compute $\mu^{\prime}$ and $\sigma^{\prime}$.  
Empirically, we found that predictions are improved if the latent variables are independent of spatial location: that is, we average-pool the outputs of these convolutional layers across all spatial locations. 
We use eight latent variables to encode the normalized trajectories across the entire image. 
At training time, we can sample the $z$ input to the decoder tower as $z=\mu^{\prime}+\eta \circ \sigma^{\prime}$ where $\eta\sim\mathcal{N}(0,1)$. $\mu^{\prime}$ and $\sigma^{\prime}$ also feed into a loss layer which computes the $\mathcal{KL}$ divergence to the $\mathcal{N}(0,1)$ distribution.
This results in a network described as: $C(96,11) \rightarrow LRN \rightarrow P \rightarrow C(256, 5) \rightarrow LRN \rightarrow P \rightarrow C(384,3) \rightarrow C(384,3) \rightarrow C(256,3) \rightarrow C(8,1) \times 2.$

\noindent {\bf Decoder Tower:}
We replicate the sampled $z$ values across spatial dimensions and multiply them with the output of the image tower with an offset.  This serves as input to four additional 256-channel convolutional layers which constitute the decoder. The fifth convolutional layer is the predicted trajectory space over the image. This can be summarized by: $C(256,3)_{1}\rightarrow C(256,3)_{2} ... \rightarrow C(256,3)_{4}  \rightarrow C(10,3)$. This output is over a coarse resolution, i.e., 16x20 pixels.   The simplest loss layer for this is the pure Euclidean loss, which corresponds to log probability according to our model (Equation~\ref{eq:model}).  However, we empirically find much faster convergence if we split this loss into two components: one is the \textit{normalized} version of the trajectory, and the other is the magnitude (with a separate magnitude for horizontal and vertical motions).  
%The decoder has two outputs both trained through euclidean loss. First, it produces trajectories normalized by the average magnitude of global image motion for both the column and row components. We empirically find this normalization essential for quick convergence. 
Because the amount of absolute motion varies considerably between images---and in particular, some action categories have much less motion overall---the normalization is required so that the learning algorithm gives equal weight to each image.  
%In order to reconstruct trajectories, the network also estimates these global magnitude constants in both directions. As these trajectories are normalized by the average magnitude of the trajectories over the whole image, the network also has a layer that estimates this magnitude. 
The total loss function is therefore:
\vspace{-0.05in}
\begin{equation}
\begin{split}
L(X,Y) = ||Y_{\text{norm}} - \hat{Y}_{\text{norm}}||^{2} + ||M_{x} - \hat{M}_{x}||^{2} + ||M_{y} - \hat{M}_{y}||^{2} \\ + \mathcal{KL}\left[Q(z|X,Y)\|\mathcal{N}(0,1)\right]
\end{split}
\end{equation}
\vspace{-0.05in}

Where $Y$ represents trajectories, $X$ is the image, $M_{i}$ are the global magnitudes, and $\hat{Y}$, $\hat{M}_{i}$ are the corresponding estimates by our network. The last term is the KL-divergence loss of the autoencoder. 
%During test time, the encoder tower is removed, and the only input to the decoder is the image tower and randomly sampled latent variables from $\mathcal{N}(0,1)$. The latent variables modify the types of predicted motion in the scene. 
We find empirically that it also helps convergence to separate both the latent variables and the decoder pathways that generate $\hat{Y}_{\text{norm}}$ from the ones that generate $\hat{M}$, perhaps due to the differences in scaling between these two outputs.

\noindent {\bf Coarse-to-Fine:}
The network as described above predicts trajectories at a stride of 16, i.e., at $1/16$ the resolution of the input image.  This is often too coarse for visualization, but training directly on higher-resolution outputs is slow and computationally wasteful.  Hence, we only begin training on higher-resolution trajectories after the network is close to convergence on lower resolution outputs.  We ultimately predict three spatial resolutions---1/16, 1/8, and 1/4 resolution---in a cascade manner similar to~\cite{Eigen15}. The decoder outputs directly to a 16x20 resolution. For additional resolution, we upsample the underlying feature map and concatenate it with the conv4 layer of the image tower. We pass this through 2 additional convolution layers, $D=C(256, 5) \rightarrow C(10, 5)$, to predict at a resolution of 32x40. Finally, we upsample this feature layer $D$, concatenate it with the conv1 layer of the image tower, and send it through one last layer of $C(10,5)$ for a final output of 64x80. 

\noindent {\bf Implementation Details:}
Given videos, we extract one-second (31-frame) clips, use~\cite{Wang13} to stabilize them respective to the first frame, and generate the trajectories which we use as a label.  As the implementation of~\cite{Wang13} tracks pixels over different scales, we take the average trajectory over all scales for a given pixel.  For each pixel in the first frame of a clip, we encode its trajectory via an $x$- and $y$-offset relative to the pixel's start location for each subsequent frame (a 60-dimensional vector).  We perform a discrete cosine transform separately for the $x$ and $y$ offsets and take only the first 5 components of each.  We use batch normalization~\cite{Ioffe15} to train the network, adding a batch normalization layer after every convolution layer that does not produce an output where scale is meaningful (i.e. $\mu,\mu^{\prime},\sigma^{\prime}$).

%% file: experiments.tex
\begin{figure*}
\centering
\begin{tabular}{ccc}
 \rotatebox{90}{\hspace{0.13in}{\bf{Prediction 1}}} &
 \includegraphics[height=0.14\textheight,width=0.3\textwidth]{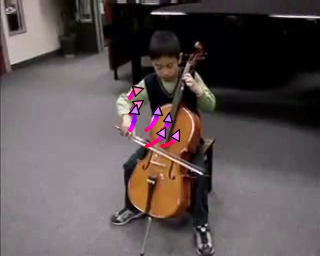} &
\includegraphics[height=0.14\textheight,width=0.6\textwidth]{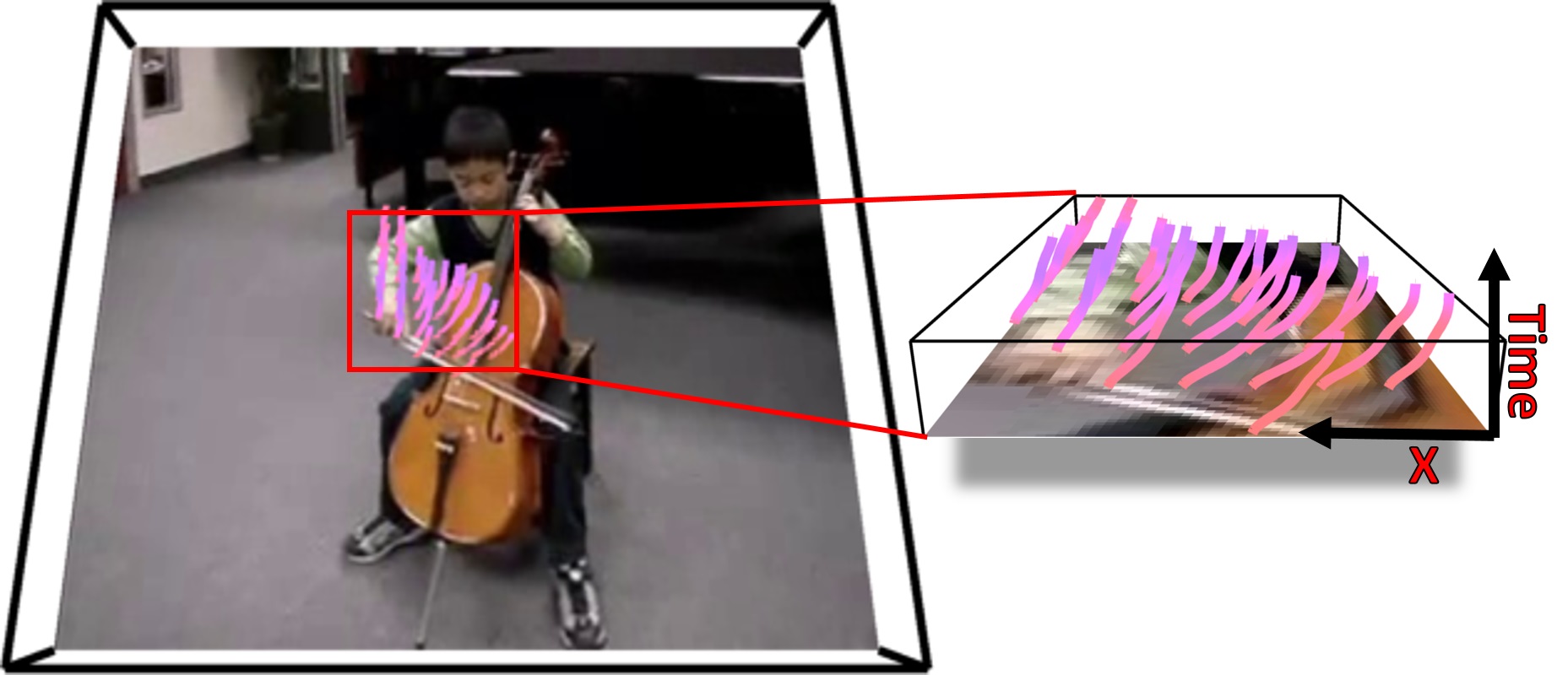} \\
 \rotatebox{90}{\hspace{0.13in}{\bf{Prediction 2}}} &
\includegraphics[height=0.14\textheight,width=0.3\textwidth]{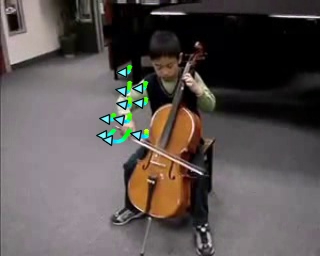} &
\includegraphics[height=0.14\textheight,width=0.6\textwidth]{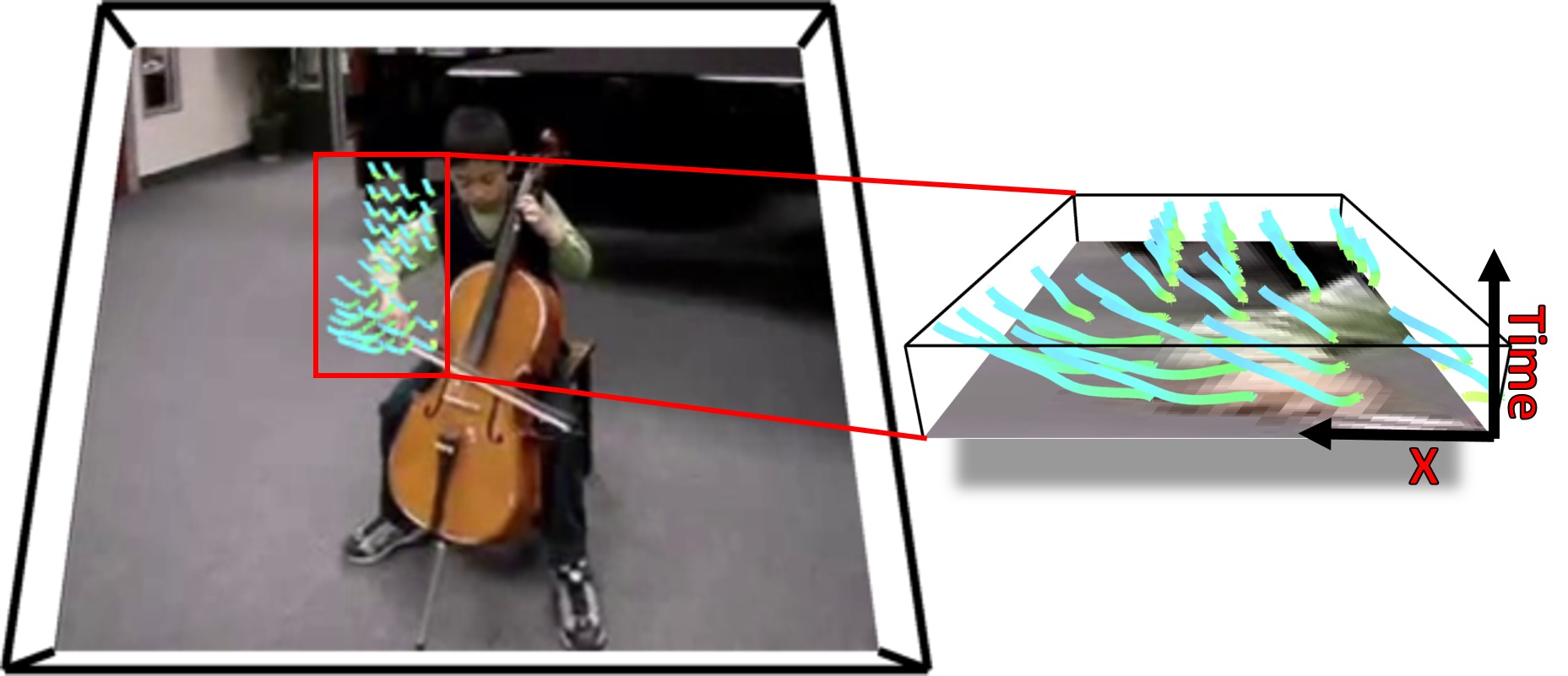} \\
 \rotatebox{90}{\hspace{0.13in}{\bf{Prediction 1}}} &
 \includegraphics[height=0.14\textheight,width=0.3\textwidth]{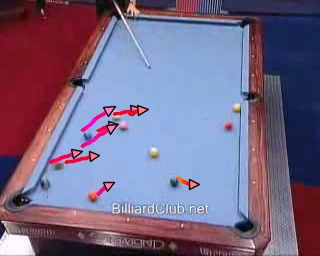} &
\includegraphics[height=0.14\textheight,width=0.6\textwidth]{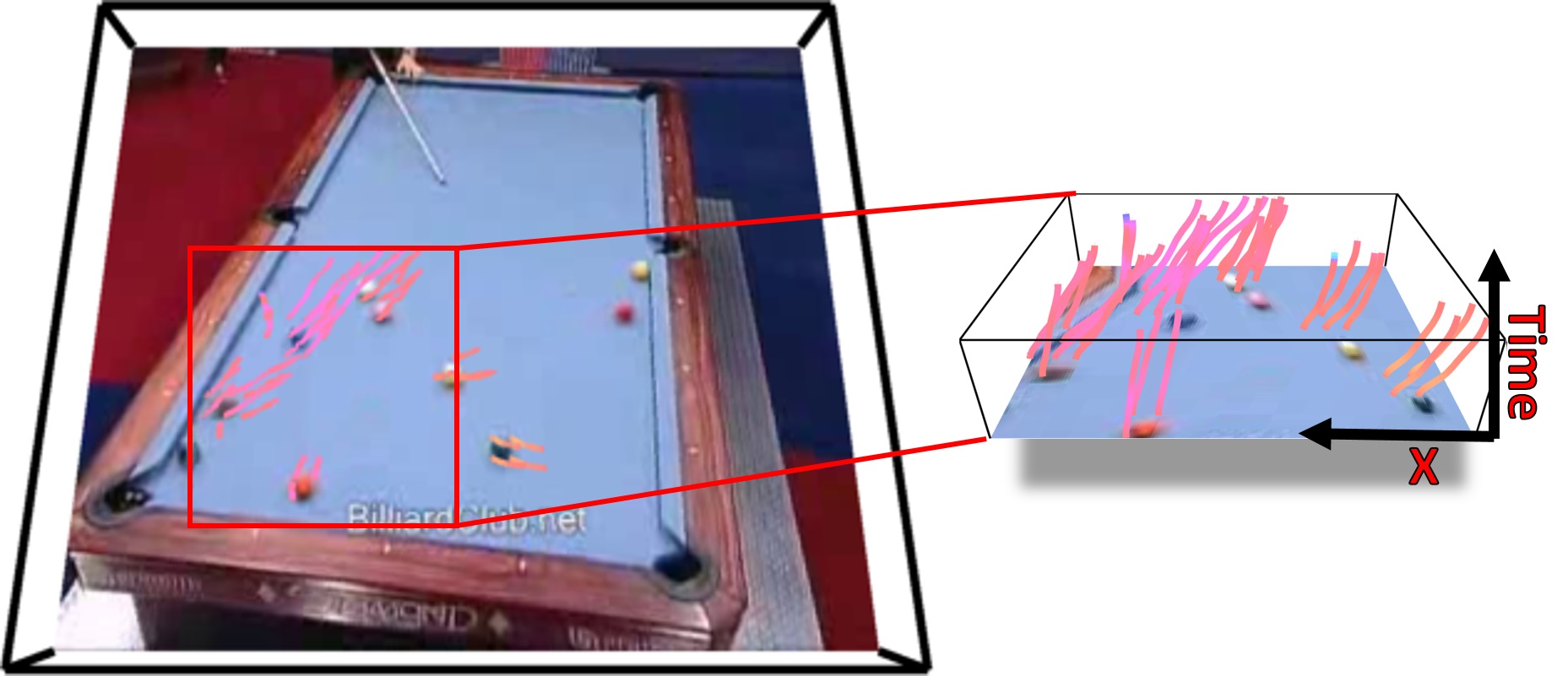} \\
 \rotatebox{90}{\hspace{0.13in}{\bf{Prediction 2}}} &
\includegraphics[height=0.14\textheight,width=0.3\textwidth]{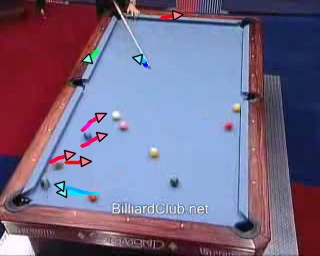} &
\includegraphics[height=0.14\textheight,width=0.6\textwidth]{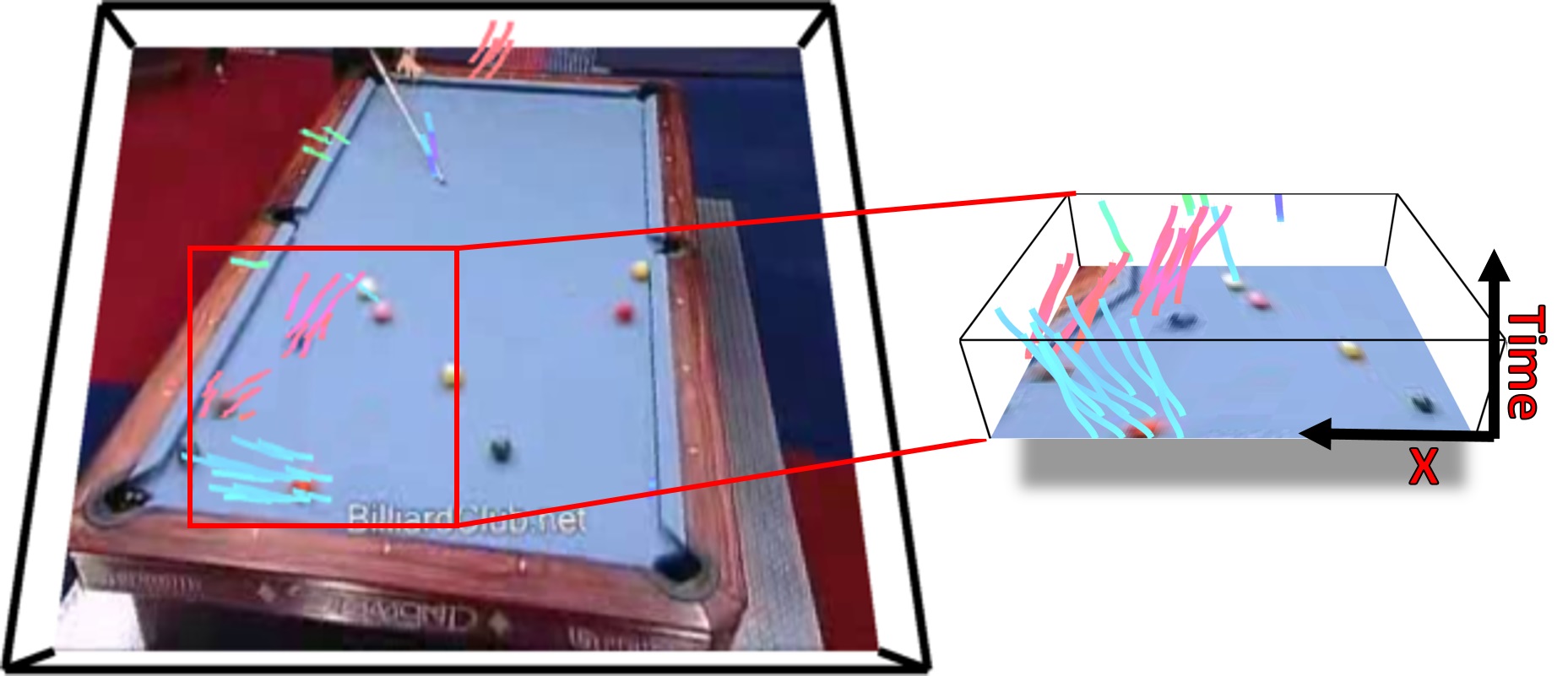} \\
 \rotatebox{90}{\hspace{0.13in}{\bf{Prediction 1}}} &
 \includegraphics[height=0.14\textheight,width=0.3\textwidth]{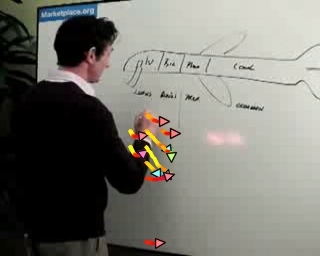} &
\includegraphics[height=0.14\textheight,width=0.6\textwidth]{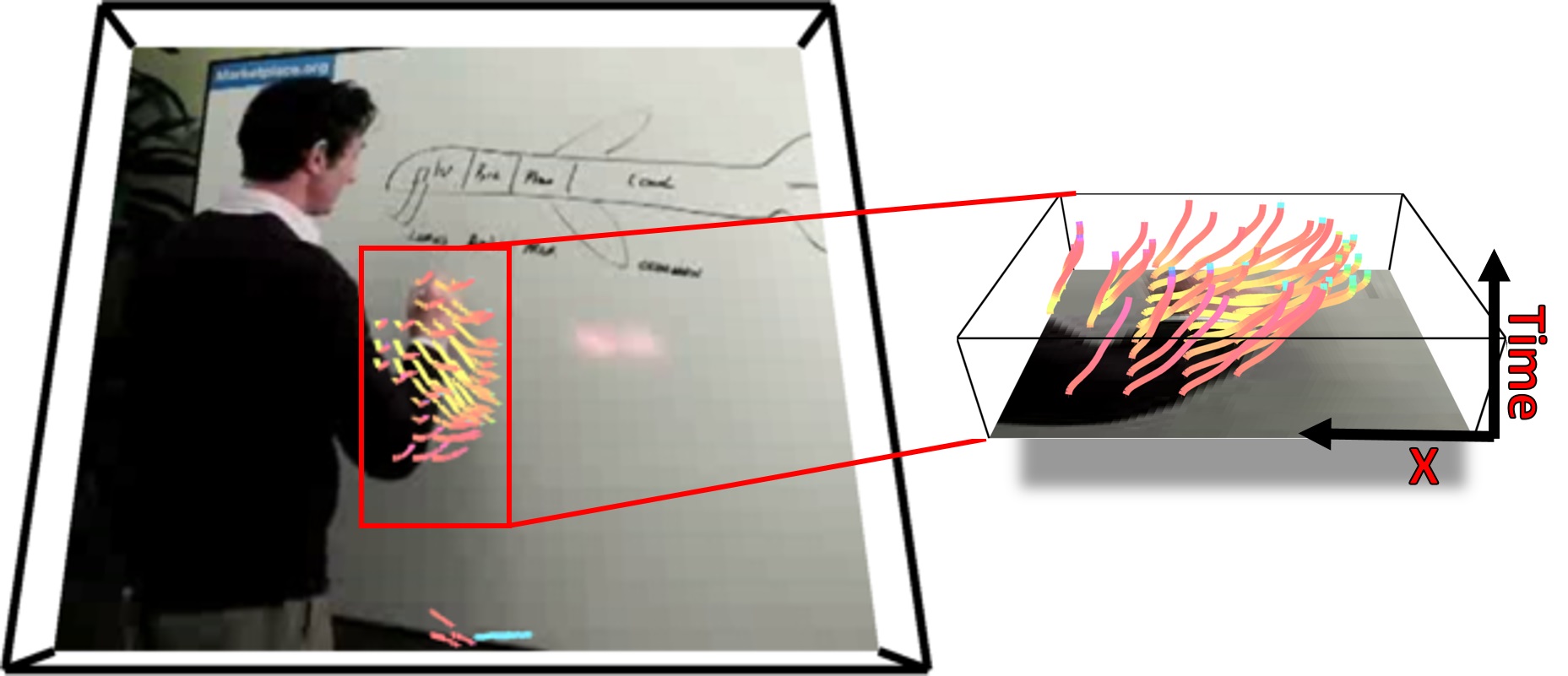} \\
 \rotatebox{90}{\hspace{0.13in}{\bf{Prediction 2}}} &
 \includegraphics[height=0.14\textheight,width=0.3\textwidth]{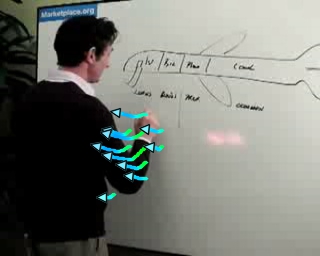} &
\includegraphics[height=0.14\textheight,width=0.6\textwidth]{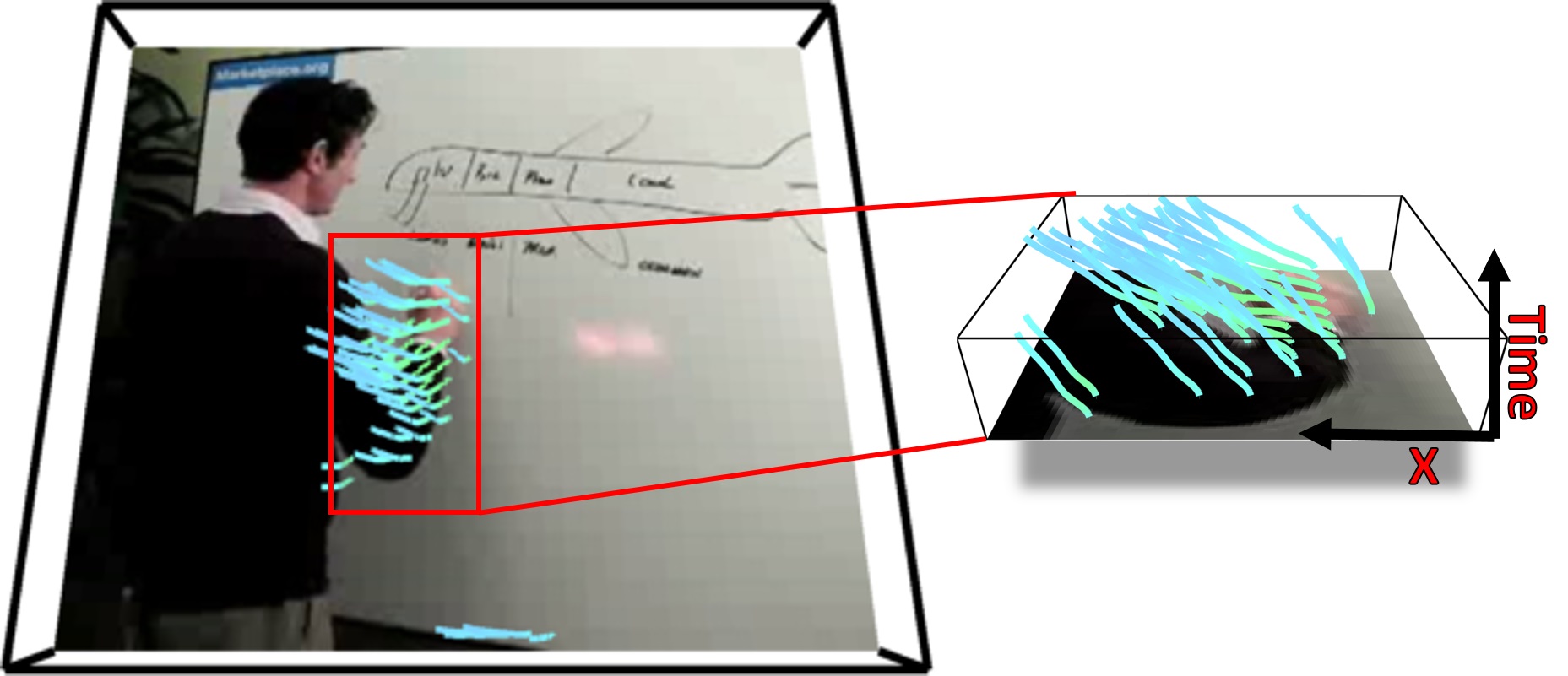} \\
 & (a) Trajectories on Image & (b) Trajectories in Space-Time
\setlength{\tabcolsep}{10pt}
\end{tabular}{}
\begin{tabular}{ p{10cm} r }
\vspace{-0.2in}
\caption{\scriptsize Predictions of our model based on clustered samples. The directions of the trajectories at each point in time are color-coded according to the square on the right. On the right is a full view of two predicted motions in 3D space-time; on the left is the projection of the trajectories onto the image plane. Best seen in our \href{http://www.cs.cmu.edu/~jcwalker/DTP/DTP.html}{videos}.}
\label{qualitative_leftright} & \raisebox{-1.0\height}{\protect\includegraphics[scale=0.30]{figures/pinwheel.jpg}}
\end{tabular}{}
\end{figure*}

\begin{figure*}
\centering
\begin{tabular}{ccc}
 \rotatebox{90}{\hspace{0.13in}{\bf{Prediction 1}}} &
 \includegraphics[height=0.14\textheight,width=0.3\textwidth]{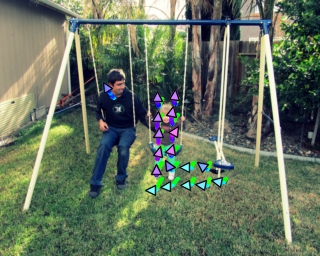} &
\includegraphics[height=0.14\textheight,width=0.6\textwidth]{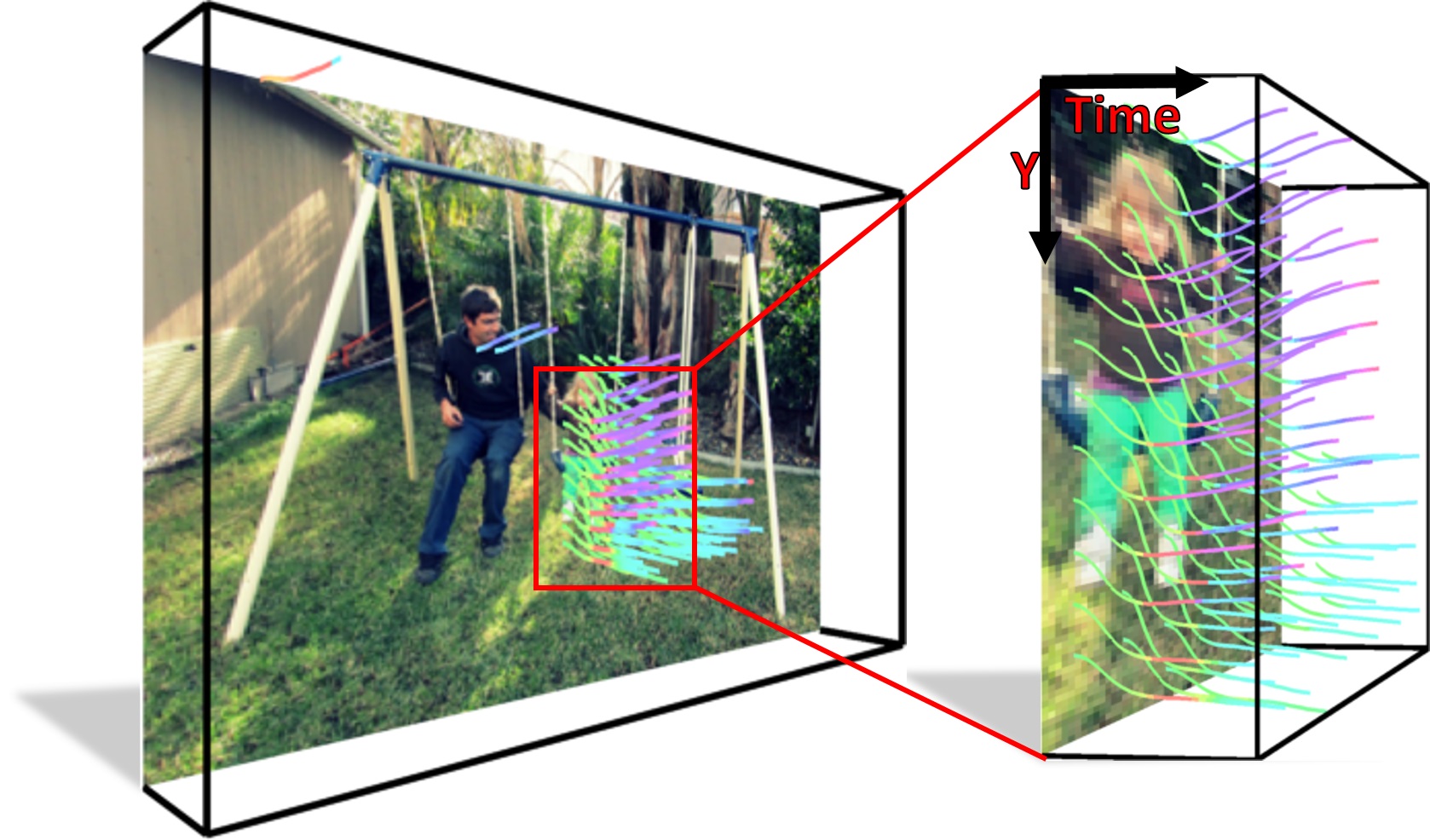} \\
 \rotatebox{90}{\hspace{0.13in}{\bf{Prediction 2}}} &
\includegraphics[height=0.14\textheight,width=0.3\textwidth]{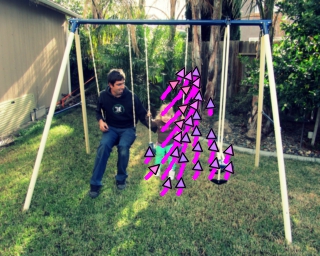} &
\includegraphics[height=0.14\textheight,width=0.6\textwidth]{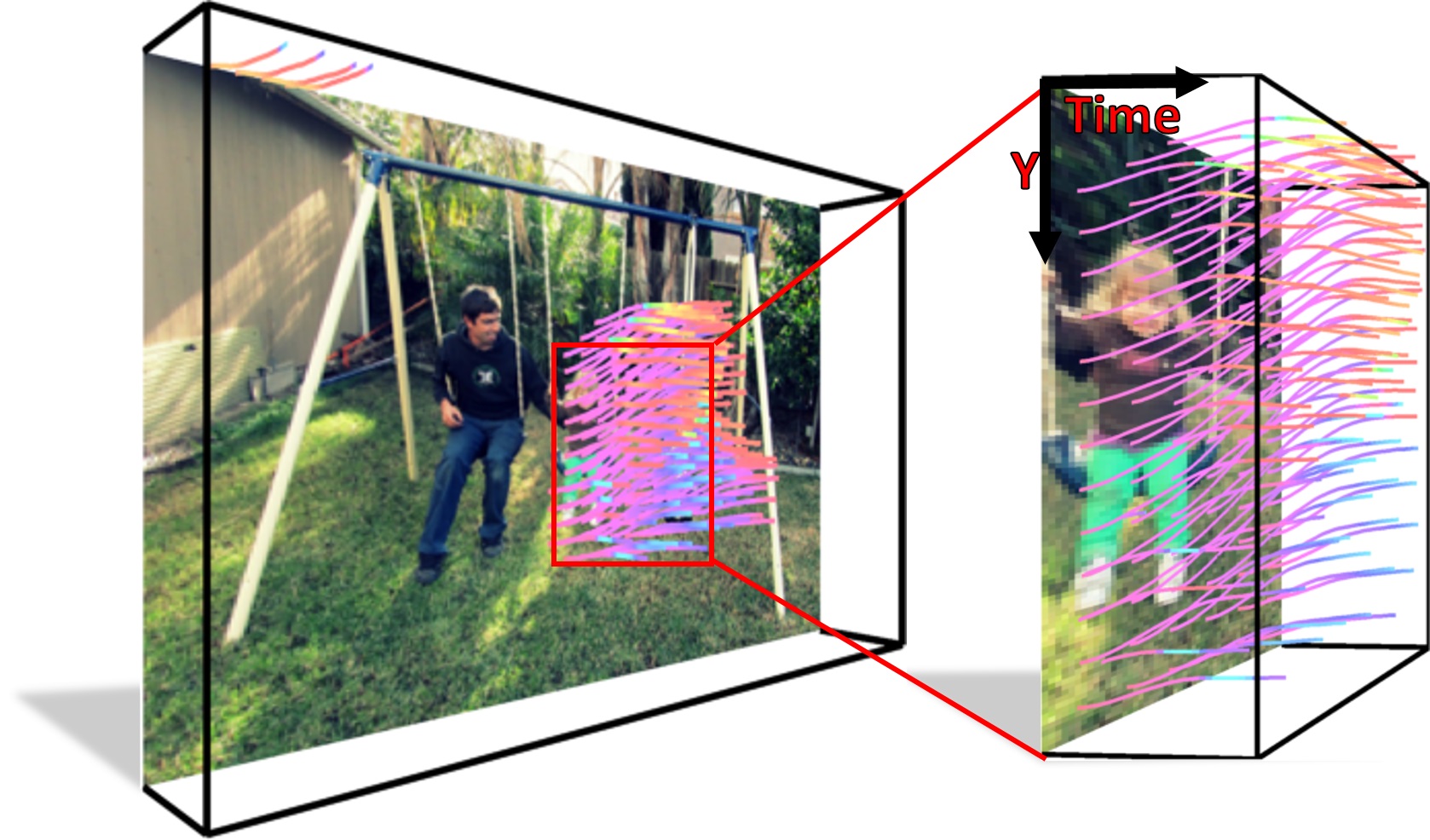} \\
 \rotatebox{90}{\hspace{0.13in}{\bf{Prediction 1}}} &
 \includegraphics[height=0.14\textheight,width=0.3\textwidth]{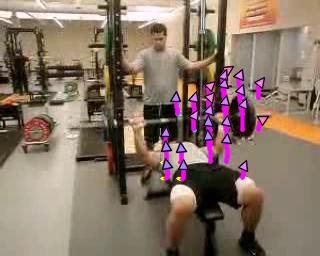} &
\includegraphics[height=0.14\textheight,width=0.6\textwidth]{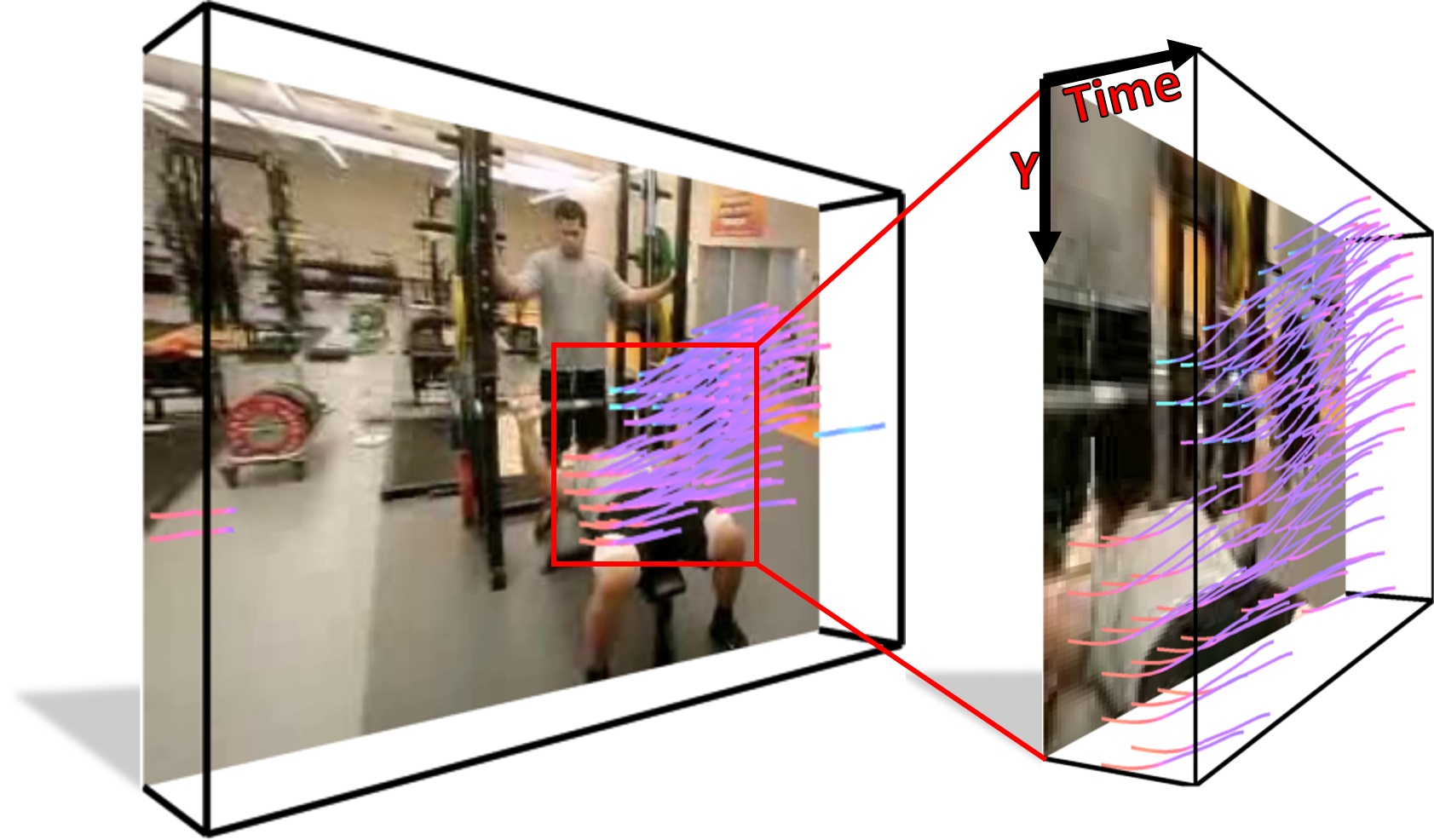} \\
 \rotatebox{90}{\hspace{0.13in}{\bf{Prediction 2}}} &
\includegraphics[height=0.14\textheight,width=0.3\textwidth]{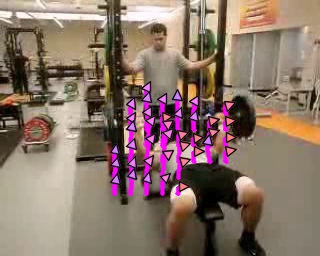} &
\includegraphics[height=0.14\textheight,width=0.6\textwidth]{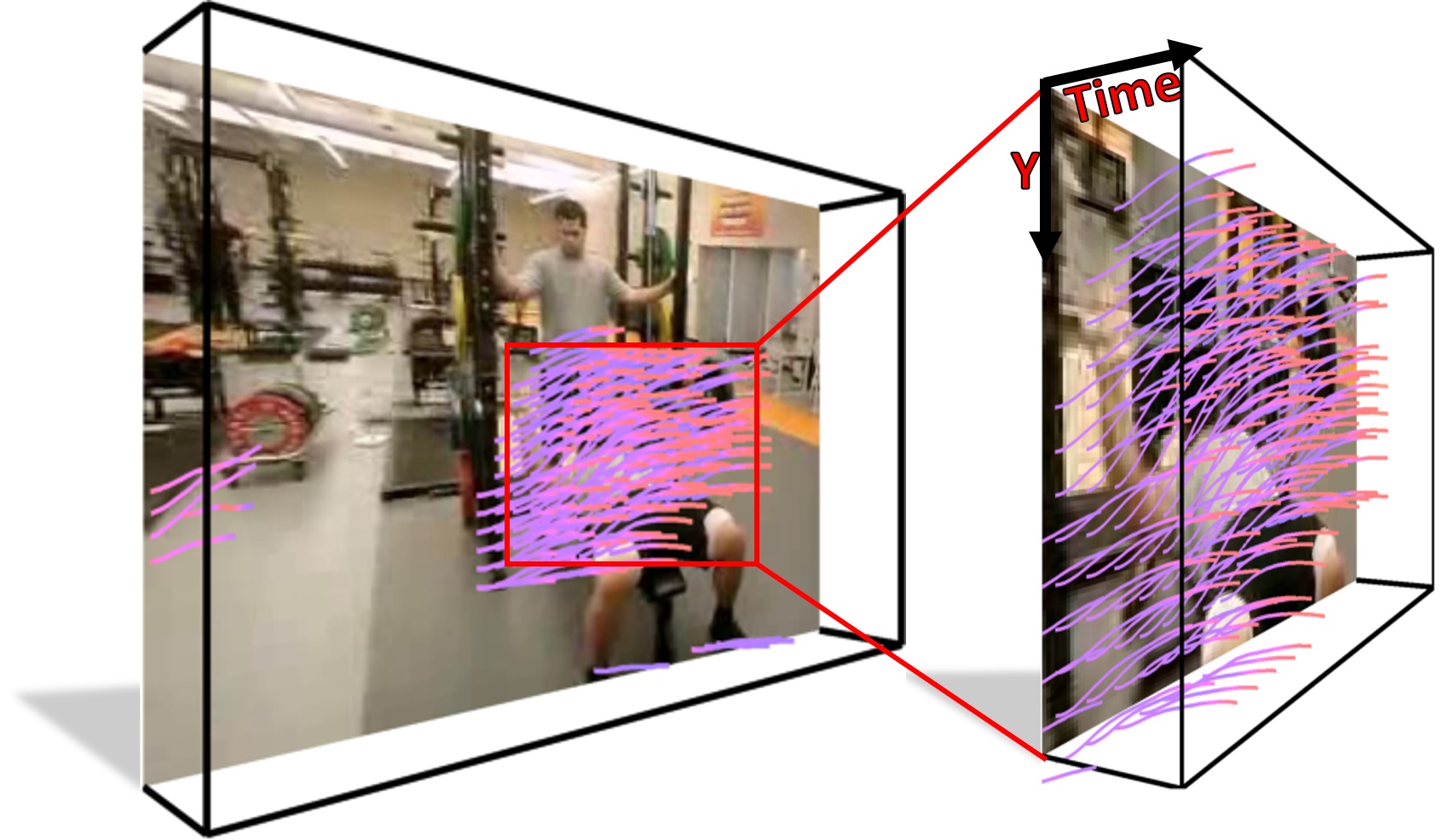} \\
 \rotatebox{90}{\hspace{0.13in}{\bf{Prediction 1}}} &
 \includegraphics[height=0.14\textheight,width=0.3\textwidth]{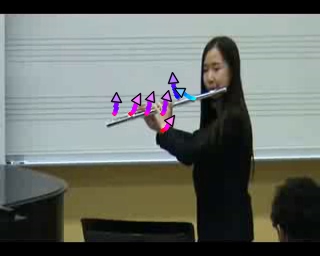} &
\includegraphics[height=0.14\textheight,width=0.6\textwidth]{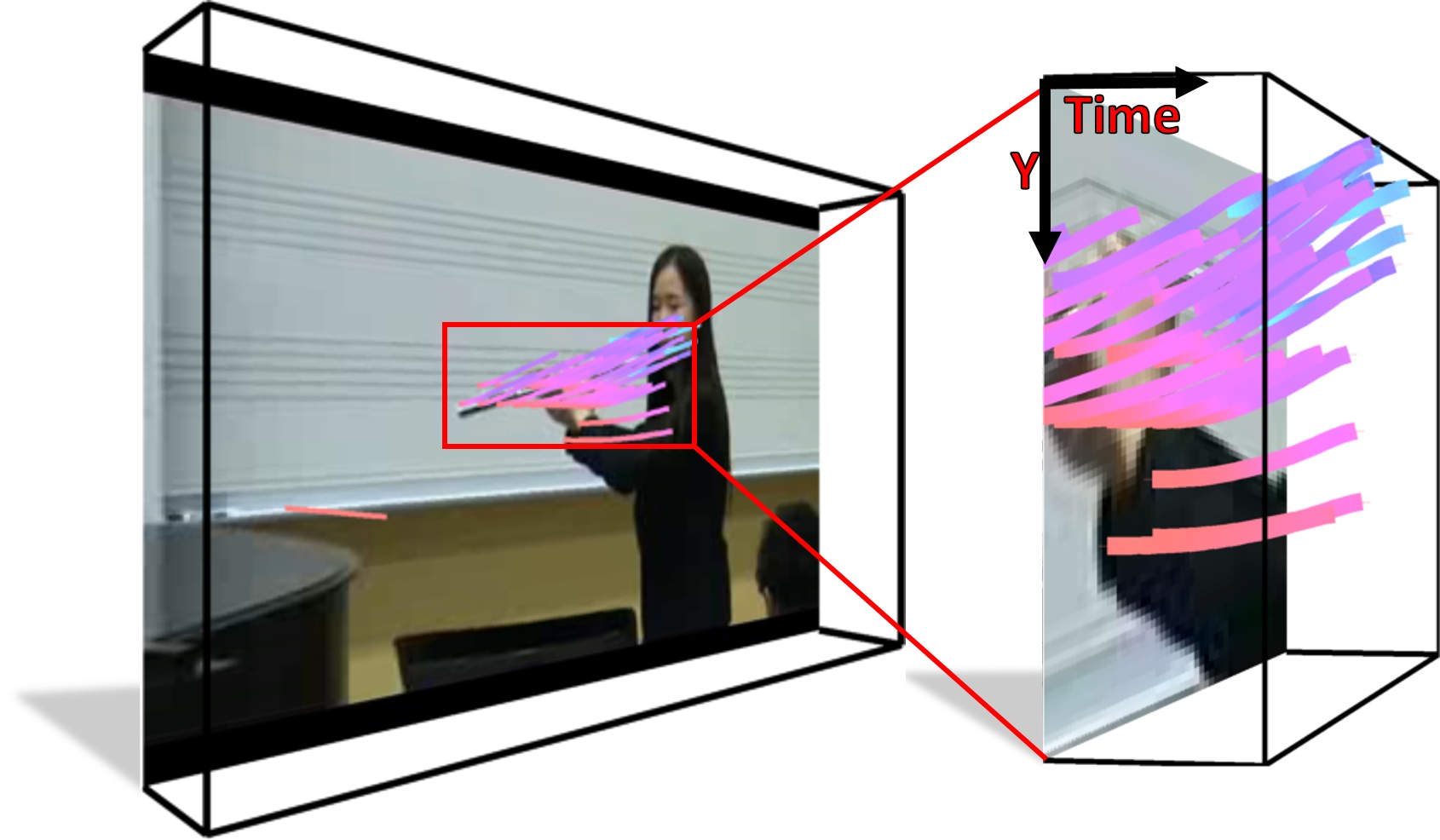} \\
 \rotatebox{90}{\hspace{0.13in}{\bf{Prediction 2}}} &
 \includegraphics[height=0.14\textheight,width=0.3\textwidth]{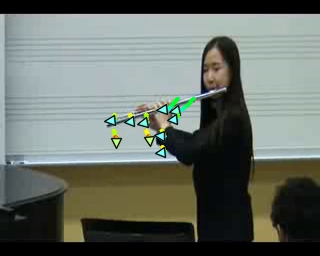} &
\includegraphics[height=0.14\textheight,width=0.6\textwidth]{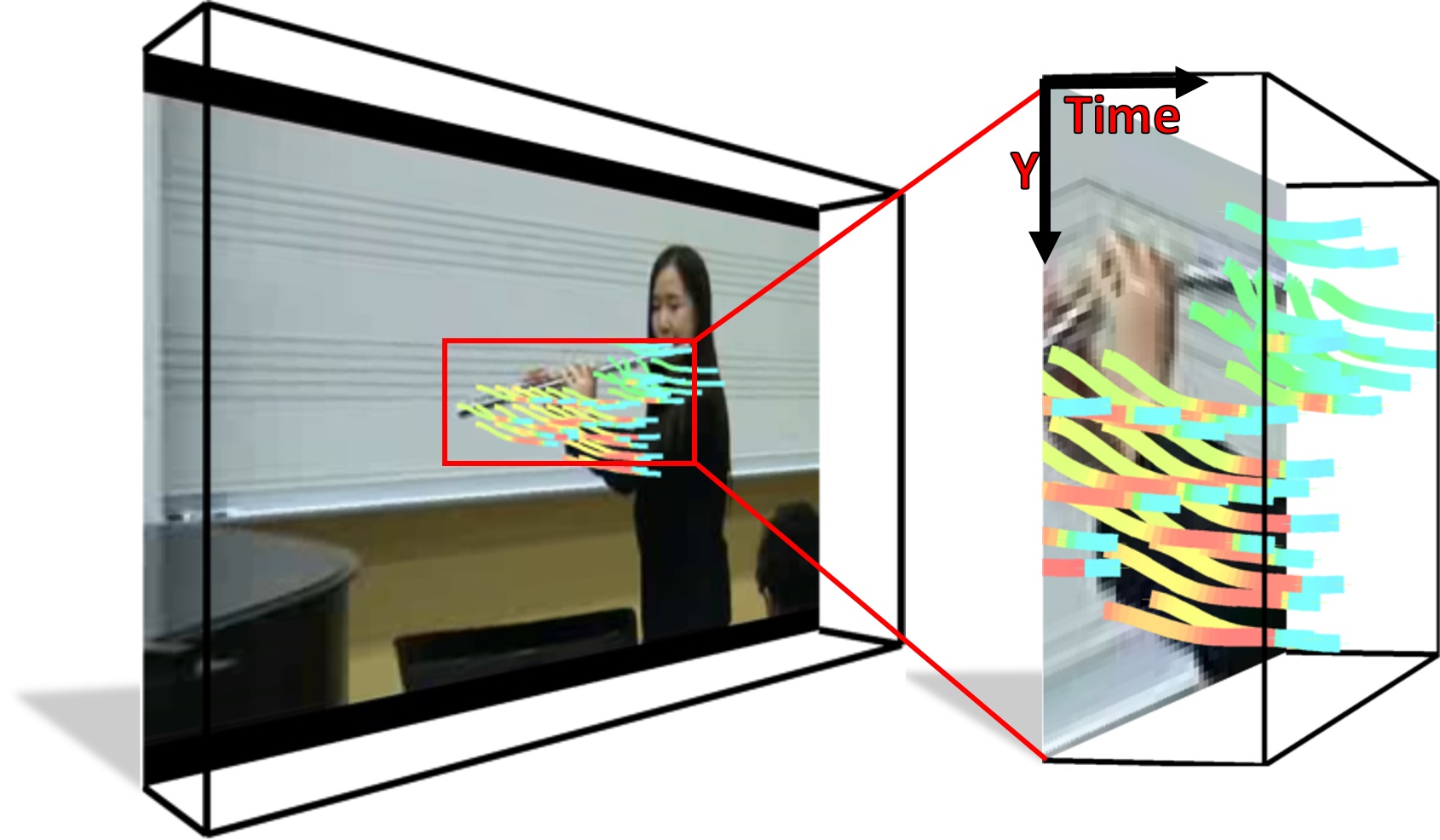} \\
 & (a) Trajectories on Image & (b) Trajectories in Space-Time
\setlength{\tabcolsep}{10pt}
\end{tabular}{}
\begin{tabular}{ p{10cm} r }
\vspace{-0.2in}
\caption{\scriptsize Predictions of our model based on clustered samples. The directions of the trajectories at each point in time are color-coded according to the square on the right. On the right is a full view of two predicted motions in 3D space-time; on the left is the projection of the trajectories onto the image plane. Best seen in our \href{http://www.cs.cmu.edu/~jcwalker/DTP/DTP.html}{videos}.}
\label{qualitative_updown} & \raisebox{-1.0\height}{\protect\includegraphics[scale=0.30]{figures/pinwheel.jpg}}
\end{tabular}{}
\end{figure*}

Because almost no prior work has focused on motion prediction beyond the timescale of optical flow, there are no established metrics or datasets for the task. 
For our quantitative evaluations, we chose to train our network on videos from the UCF101 dataset~\cite{Soomro12}. 
Although there has been much recent progress on this dataset from an action recognition standpoint, pixel-level prediction on even the UCF101 dataset has proved to be non-trivial~\cite{Srivastava15,Ranzato14}. 
Because the scene diversity is low in this dataset, we utilized as much training data as possible, i.e., all the videos except for a small hold out set for every action. 
We sampled every 3rd frame for each video, creating a training dataset of approximately 650,000 images. 
Testing data for quantitative evaluation came from the testing portion of the THUMOS 2015 challenge dataset~\cite{Gorban15}. 
The UCF101 dataset is the training dataset for the THUMOS challenge, and thus THUMOS is a relevant choice for the testing set. 
We randomly sampled 2800 frames and their corresponding trajectories for our testing data. We will make this list of frames publicly available. We use two baselines for trajectory prediction. 
The first is a direct regressor (i.e., no autoencoder) for trajectories using the same layer architecture from the image data tower. 
The second baseline is the optical flow prediction network from~\cite{Walker15}, which was trained on the same dataset. We simply extrapolate the predicted motions of the network over one second. 

%We then estimate the bandwidth for the normalized trajectories as well as the bandwidth for the magnitude. 

%As Parzen window estimates can be misleading when the dimensionality is high ~\cite{Theis15}, we we also use metrics based on Euclidean distance. As the output is multi-modal, we cluster 800 samples via k-means and then rank the clusters according to size. In this way, we can measure if a given cluster is likely to be an outlier in the set of samples. We assign the ground truth to one of the clusters. We then measure the Euclidean distance to the matched cluster and the matched cluster's rank. Because the majority of pixels do not move in the scene, we also consider a euclidean distance weighted by the absolute motion of ground truth trajectory pixels. In this way, we can measure the error specifically on the moving portions of the image. 

\begin{table}[t!]
\centering
\caption{\scriptsize Quantitative Results on the THUMOS 2015 Dataset. Lower is better.}
\begin{tabular}{|c|c|}
 \hline
Method & Negative Log Likelihood \\ \hline
Regressor & 11463 \\ \hline
Optical Flow~\cite{Walker15} & 11734 \\ \hline
Ours &~\bf{11082} \\ \hline
\end{tabular}
\vspace{-0.2in}
\label{quantitative}
\end{table}

%\begin{table}[t!]
%\centering
%\caption{\scriptsize Quantitative Results on the THUMOS 2015 Dataset. Lower is better.}
%\begin{tabular}{|c|c|}
% \hline
%Method & Negative Log Likelihood \\ \hline
%Regressor & 11463 \\ \hline
%Optical Flow~\cite{Walker15} & 11734 \\ \hline
%Ours &~\bf{11082} \\ \hline
%\end{tabular}
%\vspace{-0.2in}
%\label{quantitative}
%\end{table} 
\vspace{0.1in}
\subsection{Quantitative Results - Log Likelihood}
\vspace{-0.05in}
Choosing an effective metric for future trajectory prediction is challenging since the problem is inherently multi-modal. There might be multiple correct trajectories for every testing instance. 
Simple metrics like Euclidean distance from the ground truth become difficult to interpret in this situation: the optimal prediction for such a metric would be one which lies in-between the possibilities, a prediction which is not necessarily sensible in itself. 
We thus first evaluate our method in the context of generative models: we evaluate whether our method estimates a distribution where the ground truth is highly probable. 
Namely, given a testing example, we estimate the full conditional distribution over trajectories and calculate the log-likelihood of the ground truth trajectory under our model. 
For log-likelihood estimation, we construct Parzen window estimates using samples from our network, using a Gaussian kernel. 
In our evaluations, we compared trajectories on the coarse resolution output---$10\times16\times20$---resulting in a 3200-dimensional vector space. 
We estimate the optimal bandwidth for the Parzen window via gridsearch on the training data. 
As the networks were originally trained to optimize over normalized trajectories and magnitude separately, we also separate normalized trajectory from magnitude in the testing data, and we estimate bandwidths separately for normalized trajectories and magnitudes. 
We used 800 samples per image for Parzen window estimation. 

To evaluate the log-likelihood of the ground truth under our first baseline---the regressor---we treat the regressor's output as a mean of a multivariate Gaussian distribution.
In order to obtain an upper bound on the log-likelihood for the regressor, we optimize the bandwidths (i.e. the standard deviation parameters for the direction and magnitude, which are shared across the entire dataset) over the testing data. 
We then estimate the log-likehood of the ground-truth trajectory under this distribution. 
The optical flow network uses a soft-max layer to estimate the per-pixel distribution of motions in the image; we thus take samples of motions using this estimated distribution. 
We then use the samples to estimate a density function in the same manner as the VAE. 
In the same way for the regressor, we optimize the bandwidth over the testing data in order to obtain an upper bound for the likelihood.

In Table~\ref{quantitative}, we show our evaluations on the baselines for trajectory prediction. 
Based on the mean log-likelihood of the ground-truth trajectories under each model, our method outperforms a regressor trained on this task with the same architecture as well as extrapolation from an optical flow predictor. 
This is reasonable 
%considering that the latent variables allow our network to estimate a distribution that better fits the distribution of possible future motions. 
since the regressor is inherently unimodal: it is unable to predict distributions where there may be many reasonable futures, which Figures~\ref{qualitative_leftright} and~\ref{qualitative_updown} suggest is rather common. 
Interestingly, extrapolating the predicted optical flow 
\begin{wrapfigure}{L}{0.4\textwidth}
    \includegraphics[height=0.2\textheight,width=0.4\textwidth]{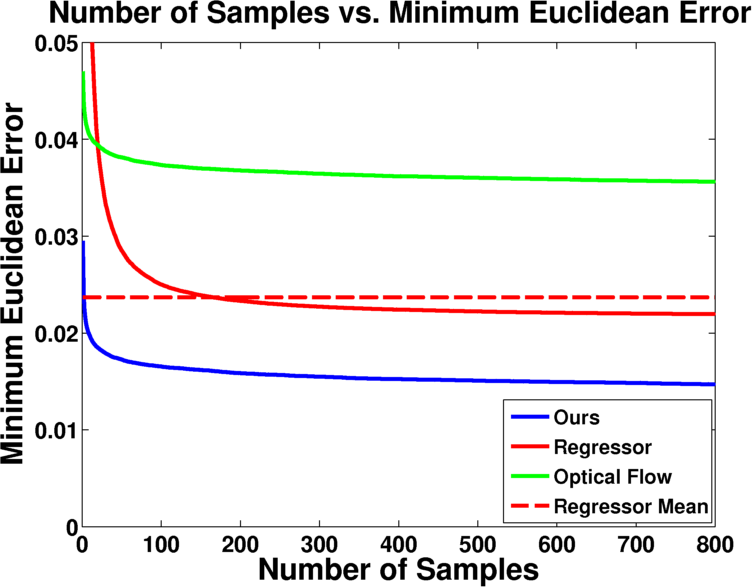}
    \caption{\footnotesize Average Minimum Euclidean distance for each method. We take the closest prediction in each testing image and plot the average of these distances as the number of samples grows per image.}
    \label{euclidean_distance}
    \vspace{-0.5in}
\end{wrapfigure}

\hspace{-0.25in} from~\cite{Walker15} does not seem to be effective, as motion may change direction considerably even over the course of one second. 

\vspace{-0.1in}
\subsection{Quantitative Results - Euclidean Distance}
\vspace{-0.05in}

As log-likelihood may be difficult to interpret, we use an additional metric for evaluation. 
While average Euclidean distance over all the samples in a particular image may not be particularly informative, it may be useful to know what was the best sample created by the algorithm. 
Specifically, given a set number $n$ of samples per image, we measure the Euclidean distance of the closest sample to the ground truth and average over all the testing images. 
For a reasonable comparison, it is necessary to make sure that every algorithm has an equal number of chances, so we take precisely $n$ samples from each algorithm per image. 
Our framework can naturally output multiple predictions. 
For the optical flow baseline~\cite{Walker15}, we can take samples from the underlying softmax probability distribution. 
For the regressor, we sample from a multivariate Gaussian centered at the regressor output and use the bandwidth parameters estimated from grid-search. 
We plot the average minimum Euclidean distance for each method in Figure~\ref{euclidean_distance}. 
We find that even with a small number of samples, our algorithm outperforms the baselines.
The additional dashed line is the result from simply using the regressor's direct output as a mean, which is equivalent to sampling with a variance of 0. 
Note that given a single sample, the regressor outperforms our method since it directly optimized the Euclidean distance at training time.
Given more than a few samples, however, ours performs better due to the multimodality of the problem.

\begin{figure*}[t!]
\centering
\newcolumntype{?}[1]{!{\vrule width #1}}
\resizebox{\linewidth}{!}{
\begin{tabular}{c?{1.0mm}cccc}
\includegraphics[height=0.10\textheight,width=0.2\textwidth]{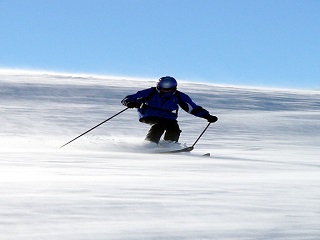} &
\includegraphics[height=0.10\textheight,width=0.2\textwidth]{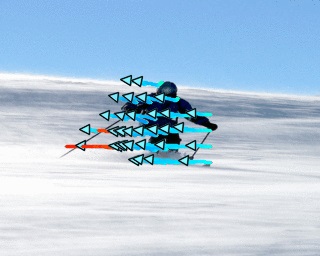} &
\includegraphics[height=0.10\textheight,width=0.2\textwidth]{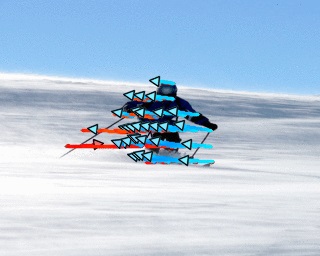} &
\includegraphics[height=0.10\textheight,width=0.2\textwidth]{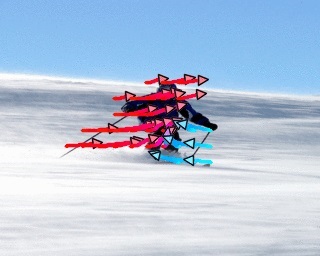} &
\includegraphics[height=0.10\textheight,width=0.2\textwidth]{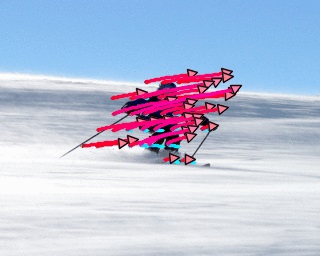} \\
\includegraphics[height=0.10\textheight,width=0.2\textwidth]{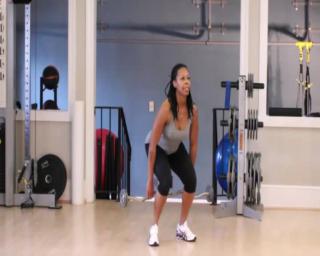} &
\includegraphics[height=0.10\textheight,width=0.2\textwidth]{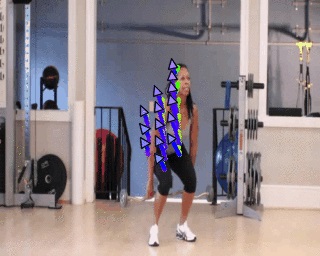} &
\includegraphics[height=0.10\textheight,width=0.2\textwidth]{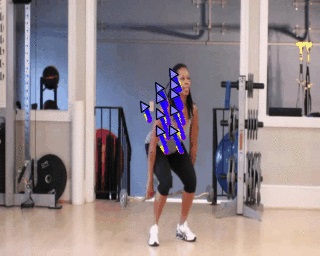} &
\includegraphics[height=0.10\textheight,width=0.2\textwidth]{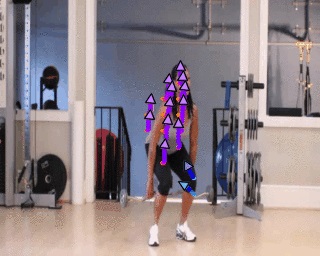} &
\includegraphics[height=0.10\textheight,width=0.2\textwidth]{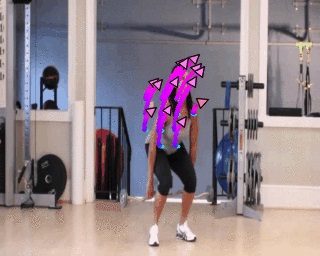} \\
\includegraphics[height=0.10\textheight,width=0.2\textwidth]{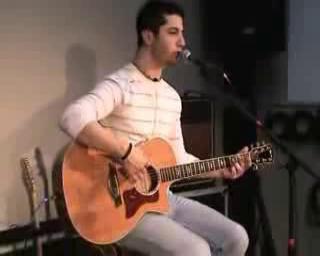} &
\includegraphics[height=0.10\textheight,width=0.2\textwidth]{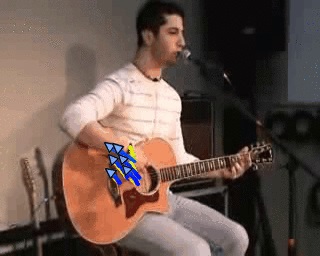} &
\includegraphics[height=0.10\textheight,width=0.2\textwidth]{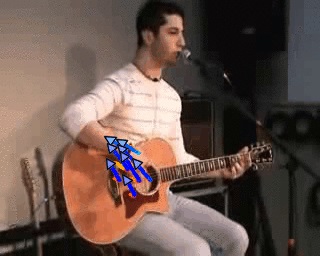} &
\includegraphics[height=0.10\textheight,width=0.2\textwidth]{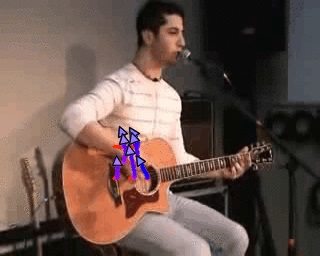} &
\includegraphics[height=0.10\textheight,width=0.2\textwidth]{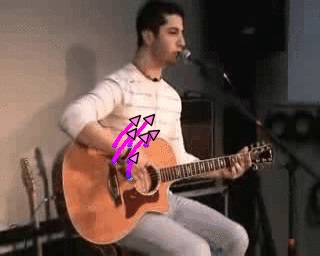} \\
\includegraphics[height=0.10\textheight,width=0.2\textwidth]{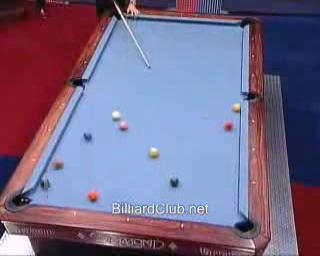} &
\includegraphics[height=0.10\textheight,width=0.2\textwidth]{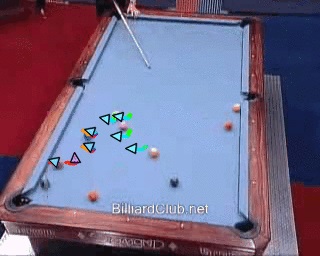} &
\includegraphics[height=0.10\textheight,width=0.2\textwidth]{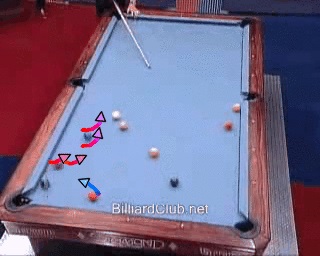} &
\includegraphics[height=0.10\textheight,width=0.2\textwidth]{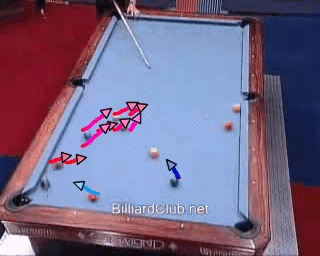} &
\includegraphics[height=0.10\textheight,width=0.2\textwidth]{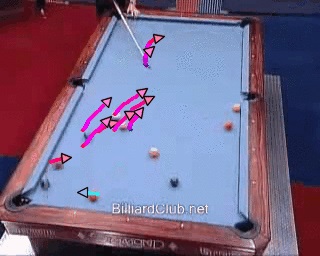} \\
\includegraphics[height=0.10\textheight,width=0.2\textwidth]{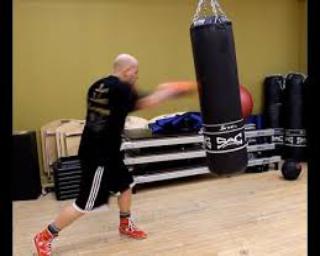} &
\includegraphics[height=0.10\textheight,width=0.2\textwidth]{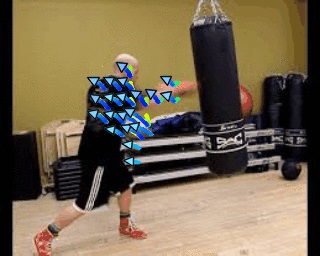} &
\includegraphics[height=0.10\textheight,width=0.2\textwidth]{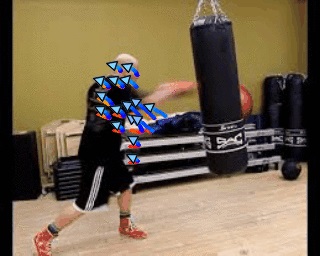} &
\includegraphics[height=0.10\textheight,width=0.2\textwidth]{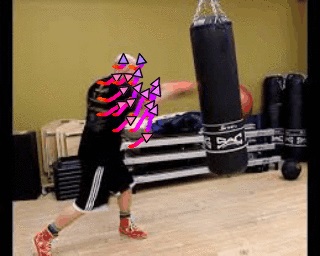} &
\includegraphics[height=0.10\textheight,width=0.2\textwidth]{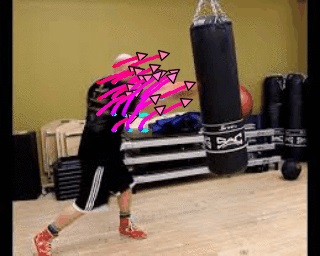} \\ 
 \bf{Input Image} & \multicolumn{4}{c}{\bf{Interpolation}}
\setlength{\tabcolsep}{10pt}
\end{tabular}{}
}
\begin{tabular}{ p{10cm} r }
\vspace{-0.2in}
\caption{Interpolation in latent variable space between two points from left to right. Each column represents a set of images with the same latent variables. Left to right represents a linear interpolation between two points in z-space. The latent variables influence direction to some extent, but the context of the image either amplifies or greatly reduces this direction. The squatting woman is always moving upward, but the skier changes drastically in direction. }
\label{interpolation} & \raisebox{-1.0\height}{\protect\includegraphics[scale=0.70]{figures/pinwheel.jpg}}
\vspace{-0.2in}
\end{tabular}{}
\end{figure*}

\vspace{-0.1in}
\subsection{Qualitative Results} 
\vspace{-0.05in}

We show some qualitative results in Figures~\ref{qualitative_leftright} and \ref{qualitative_updown}.
For these results, we cluster 800 samples into 10 clusters via Kmeans and show top two clusters with significant motion. 
Our method is able to identify active objects in the scene whether they are hands, entire bodies or objects such as billiard balls.  
The network then predicts motion based on the the context of the scene. 
For instance, the network tends to predict up and down motions for the man lifting a weight and the people on the swing in Figure~\ref{qualitative_updown}. 
The boy playing the violin moves his arm left and right, and the man writing on the board moves his arm across the board. 
%Indeed, Figures~\ref{qualitative_leftright} and \ref{qualitative_updown} show reasonable predictions given the image. 
Figure~\ref{interpolation} shows the role latent variables play in predicting motion in some selected scenes with a distinct action: interpolating between latent variable values interpolates between the motion prediction. 
Based on this figure, at least some latent variables encode the direction of motion. However, the network still depends on image information to restrict the types of motions that can occur in a scene. 
Given a set of the exact same latent variables, the predicted motion is modulated based on the action in the scene. 
For instance, the man skiing moves only left or right, while the woman squatting largely moves only up or down with small changes to the x-axis. 

Our  \href{http://www.cs.cmu.edu/~jcwalker/DTP/DTP.html}{website} includes videos of pixel motion.  We encourage our readers to look at them, since they are easier to interpret than these static visualizations. 

\vspace{-0.15in}
\subsection{Representation Learning} 
\vspace{-0.05in}
Prediction implicitly depends on a number of fundamental vision tasks: for example, the network must recognize the scene to infer the action that is being performed and must detect, localize, and infer the pose of humans and objects that may move.  
%. The given action may differ based on the context. People are likely to behave differently in a gym setting versus an office or residential environment. Second, it must perform some form of detection and localization. It must detect humans or other active objects in the scene, and then it must also decide what should move in the scene. 
Hence, we expect the representation learned for the task of motion prediction may generalize for other vision tasks. 
We thus evaluate the representation learned by our network on the task of object detection. 
We take layers from the image tower and fine-tune them on the PASCAL 2012 training dataset. 
For all methods, we apply the between-layer scale adjustment~\cite{Krahenbuhl16} to calibrate the pre-trained networks, as it improves the finetuning behavior of all methods except one.
We then compare detection scores against other unsupervised methods of representation learning using Fast-RCNN~\cite{Girshick15}. 
%We also show that our network learns a representation that transfers to different vision tasks. 
We find that from a relatively small amount of data, our method outperforms other methods that were trained on datasets with far larger diversity in scenes and types of objects. 
Interestingly, our method outperforms all unsupervised methods, even~\cite{Doersch15}, on human detection, likely because most of the movement our algorithm needs to predict comes from humans.
%of our predicted movements are from humans, this is not surprising. 
%However, we have achieved this benchmark by exposing the network to much less data compared to the other networks.

\begin{table*}[t!]
\centering
\caption{{\footnotesize mean Average Precision (mAP) on VOC 2012. The ``External data'' column represents the amount of data exposed outside of the VOC 2012 training set. ``cal'' denotes the between-layer scale adjustment~\cite{Krahenbuhl16} calibration.}}
\vspace{-0.1in}
\renewcommand{\arraystretch}{1.2}
\renewcommand{\tabcolsep}{1.2mm}
\resizebox{\linewidth}{!}{
\begin{tabular}{@{}l|l|r*{19}{c}|c@{}}
\hline
\textbf{VOC 2012 test}   &external data & aero      & bike      & bird      & boat      & bottle     & bus        & car        & cat        & chair      & cow        & table      & dog        & horse      & mbike      & person     & plant      & sheep      & sofa       & train      & tv         & mAP       \\
\hline
scratch+cal  & N/A  &  67.5  &  49.8  &  27.9  &  23.9  &  13.6  &  57.8  &  48.1  &  51.7  &  16.1  &  33.2  &  29.2 &  45.3   &  51.9  &  58.8  &  51.7  &  16.8  &  39.7  &  29.4  &  55.7  &  43.5  &  40.6  \\
kmeans~\cite{Krahenbuhl16}+cal  & N/A  &  71.1  &  56.8  &  31.8  &  28.1  &  17.7  &  62.5  &  56.6  &  59.9  &  19.9  &  37.3  &  36.2 &  52.9   &  56.4  &  64.3  &  57.1  &  21.2  &  45.8  &  39.1  &  60.9  &  46.0  &  46.1  \\
rel. pos.~\cite{Doersch15}+cal  & 1.2M ImageNet   &  \bf{74.3}  &  \bf{64.7}  &  \bf{42.6}  &  \bf{32.6}  &  25.9  &  \bf{66.5}  &  \bf{60.2}  &  \bf{67.9}  & \bf{27.0}  &  \bf{47.9}  &  \bf{41.3} &  \bf{64.5}   &  \bf{63.4}  & \bf{69.1} & 57.5  &  25.3  &  51.9  &  \bf{46.7}  &  \bf{64.6}  &  51.4  &  \bf{52.3}  \\
egomotion~\cite{Agrawal15}+cal  &20.5K KITTI img.  &  70.7  &  56.3  &  31.9  &  25.6  &  18.7  &  60.4  &  54.1  &  57.6  & 19.8  &  40.9  &  31.8 &  51.9   &  54.9  &  61.7  &  53.5  &  19.8  &  45.2  &  36.3 &  56.9  &  49.1  &  44.9  \\
vid. embed~\cite{Wang15} & 1.5M vid. frames (100k vid) &  68.8  &  62.1  &  34.7  &  25.3  &  \bf{26.6}  &  57.7  &  59.6  &  56.3  &  22.0  &  42.6  &  33.8 &  52.3   &  50.3  &  65.6  &  53.9  &  \bf{25.8}  &  51.5  &  32.3  &  51.7  &  51.8  &  46.2  \\
vid. embed~\cite{Wang15} & 5M vid. frames (100k vid) & 69.0 & 64.0 & 37.1 & 23.6 & 24.6 & 58.7 & 58.9 & 59.6 & 22.3 & 46.0 & 35.1 & 53.3 & 53.7 & 66.9 & 54.1 & 25.4 & \bf{52.9} & 31.2 & 51.9 & 51.8 & 47.0  \\
vid. embed~\cite{Wang15} & 8M vid. frames (100k vid) & 67.6 & 63.4 & 37.3 & 27.6 & 24.0 & 58.7 & 59.9 & 59.5 & 23.7 & 46.3 & 37.6 & 54.8 & 54.7 & 66.4 & 54.8 & \bf{25.8} & 52.5 & 31.2 & 52.6 & \bf{52.6} & 47.5  \\
vid. embed~\cite{Wang15}+cal  & 8M vid. frames (100k vid)  &  68.1  &  53.1  &  31.9  &  24.3  &  16.9  &  57.2  &  50.8  &  58.4  & 14.1  &  36.9  &  27.6 &  52.5   &  49.6  &  60.0  &  48.4  &  15.8  &  41.9  &  34.4  &  55.6  &  45.6  &  42.2  \\
ours+cal  & 13k UCF101 vid.  &  71.7  &  60.4  &  34.0  &  27.8  &  18.6  &  63.5  &  56.6  &  61.1  & 21.2  &  39.3  &  35.1 &  57.1   &  58.6  &  66.0  &  \bf{58.4}  &  20.5  &  45.6  &  38.3  &  62.1  &  49.9  &  47.3  \\
\hline
\end{tabular}
}
\label{tab:VOCAP}
\vspace{-0.2in}
\end{table*}

%% file: conclusion.tex
In this paper we have presented an approach to predict dense trajectories from pixels. Specifically, our framework proposes a Variational Autoencoder conditioned on images: the framework uses latent variables (and the predicted distribution) to represent multiple possible trajectories. Our method requires no human labels and can be trained efficiently with back-propagation. Furthermore, we showed that our method learns a representation that transfers to other vision tasks such as object detection. We believe future work on similar algorithms may lead to applications in graphics such as generating videos from single images. 
\vspace{0.1in}

\noindent {\bf Acknowledgements:} \small We thank the NVIDIA Corporation for the donation of Tesla K40 GPUs for this research.  In addition, this work was supported by NSF grant IIS1227495.

%We show that our method is able to predict a distribution of high-dimensional continuous motion based on the image alone. 

%Another direction is sequential prediction of motion trajectories into a longer future time span.  